\documentclass{article}
\usepackage{textcomp}
\usepackage{changes}

\usepackage{PRIMEarxiv}
\usepackage{amssymb}  
\usepackage[utf8]{inputenc} 
\usepackage[T1]{fontenc}    
\usepackage{hyperref}       
\usepackage{url}            
\usepackage{booktabs}    
\usepackage{multirow}  
\usepackage{booktabs}  
\usepackage{float}  
\usepackage{tabularx}  
\usepackage{amsfonts}       
\usepackage{nicefrac}       
\usepackage{microtype}      
\usepackage{lipsum}
\usepackage{fancyhdr}       
\usepackage{graphicx}       
\graphicspath{{media/}}     
\usepackage{siunitx}
\usepackage{amsmath}
\usepackage{tikz}
\usetikzlibrary{shapes.geometric, arrows}
\usetikzlibrary{positioning, arrows.meta, shapes.geometric}
\usepackage{graphicx}
\usepackage{caption} 
\usepackage{graphicx}
\usepackage{subcaption} 
\title{TDE-3: An improved prior for optical flow computation in spiking neural networks}

\author{
    Matthew Yedutenko\thanks{Corresponding author} \\
    Micro Air Vehicle Laboratory \\ 
    Delft University of Technology \\
    The Netherlands \\
    \texttt{m.yedutenko@tudelft.nl} \\
    \And
    Federico Paredes-Vallés \\
    Stuttgart Laboratory 1 \\
    Sony Semiconductor Solutions Europe \\
    Sony Europe B.V.\\
    \texttt{federico.paredes-valles@sony.com} \\
    \And
    Lyes Khacef \\
    Stuttgart Laboratory 1 \\
    Sony Semiconductor Solutions Europe \\
    Sony Europe B.V. \\
    \texttt{lyes.khacef@sony.com} \\
    \And
    Guido C.H.E. De Croon \\
    Micro Air Vehicle Laboratory \\
    Delft University of Technology \\
    The Netherlands \\
    \texttt{g.c.h.e.decroon@tudelft.nl} \\
}

\begin{document}
\maketitle
\begin{abstract}

Motion detection is a primary task required for robotic systems to perceive and navigate in their environment. Proposed in the literature bioinspired neuromorphic Time-Difference Encoder (TDE-2) combines event-based sensors and processors with spiking neural networks to provide real-time and energy-efficient motion detection through extracting temporal correlations between two points in space. However, on the algorithmic level, this design leads to a loss of direction-selectivity of individual TDEs in textured environments. Here we propose an augmented 3-point TDE (TDE-3) with additional inhibitory input that makes TDE-3 direction-selectivity robust in textured environments. We developed a procedure to train the new TDE-3 using backpropagation through time and surrogate gradients to linearly map input velocities into an output spike count or an Inter-Spike Interval (ISI).  Using synthetic data we compared training and inference with spike count and ISI with respect to changes in stimuli dynamic range, spatial frequency, and level of noise. ISI turns out to be more robust towards variation in spatial frequency, whereas the spike count is a more reliable training signal in the presence of noise.  We performed in-depth quantitative investigation of optical flow coding with TDE and compared TDE-2 vs TDE-3 in terms of energy-efficiency and coding precision. Results show that on the network level, both detectors show similar precision (20\textdegree angular error, 88\% correlation with ground truth). Yet, due to the more robust direction-selectivity of individual TDEs, TDE-3 –based network spike less and hence is more energy-efficient.  Reported precision is on par with model-based methods but the spike-based processing of the TDEs provides allows more energy-efficient inference with neuromorphic hardware. 

\end{abstract}
\tikzstyle{startstop} = [rectangle,
minimum width=2.5cm, 
minimum height=1cm,
text centered, 
draw=black, 
fill=orange!40]

\tikzstyle{io} = [rectangle, rounded corners, 
minimum width=2.5cm, 
minimum height=1cm, text centered, 
draw=black, 
fill=gray!80]

\tikzstyle{process} = [rectangle, 
minimum width=2.5cm, 
minimum height=1cm, 
text centered, 
draw=black, 
fill=blue!30]

\tikzstyle{null} = [rectangle, 
minimum width=2.5cm, 
minimum height=1cm, 
text centered, 
draw=white, 
fill=white!40]

\tikzstyle{decision} = [diamond, 
minimum width=2.5cm, 
minimum height=1cm, 
text-centered, 
draw=black, 
fill=blue!30]
\tikzstyle{arrow} = [thick,->,>=stealth]



\section{Introduction}
\begin{figure}[h]
\centering
\includegraphics[width=0.9\linewidth]{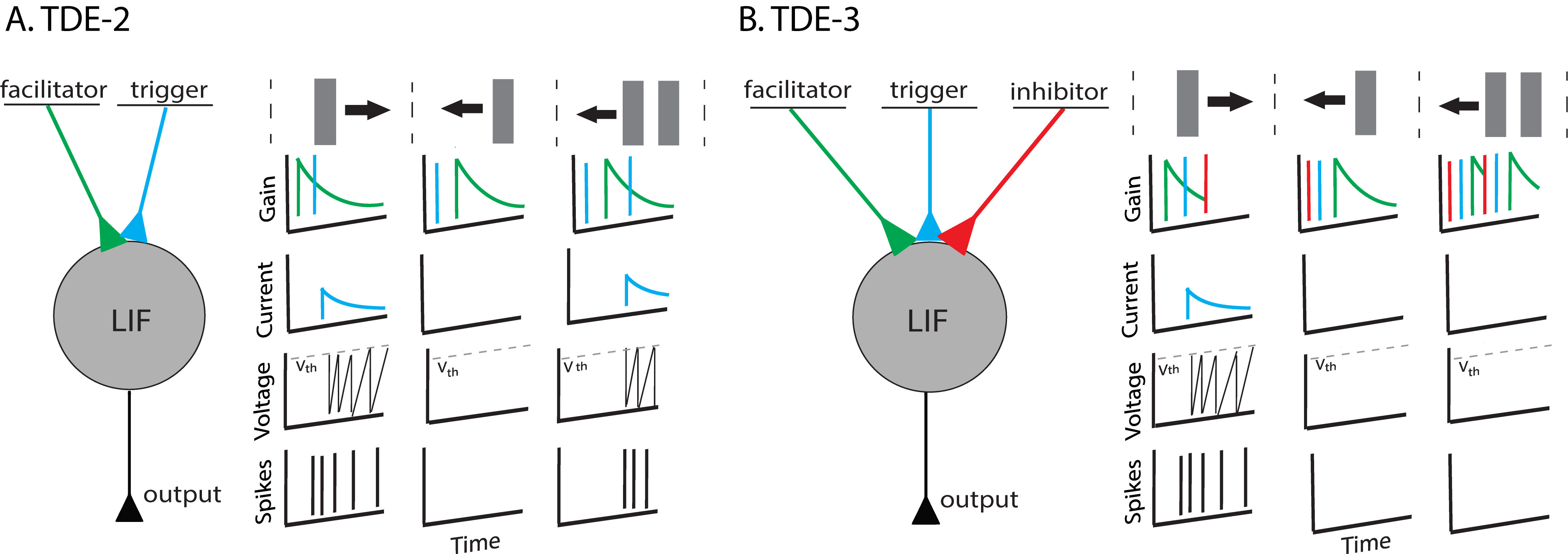}
\caption{TDE-2 and TDE-3. A. Left-to-right tuned 2-point TDE. It has two compartments: the facilitator and the trigger. When a stimulus moves to the right the trigger is activated after the facilitator and the neuron fires. However, in the gain, there is residual activity. Thus, when multiple textures move left (or orthogonally) the detector loses direction-selectivity B. Left-to-right tuned 3-point TDE. It has three compartments. Input to the inhibitor resets the gain to zero and removes residual activity. Therefore, direction-selectivity is retained.}
\label{fig:Figure1}
\end{figure}

Event-based cameras are the dawn of the brave new world of ultra-efficient neuromorphic hardware. Taking inspiration from biological retinas, these cameras transmit only temporal changes in brightness using asynchronous events \cite{Lichtsteiner2008}. As a result, they require >$10^2 \times$ less energy and memory than traditional frame-based cameras while having $10^4 \times$ higher dynamic range and up to $10^4 \times$ shorter latency \cite{Gallego_etal22}, which is particularly handy in robotics applications. Under constant illumination, temporal changes in brightness are generated by motion, therefore optical flow estimation is one of, if not the main task in event-based vision. 

The efficiency of event-based processing in the estimation of optical flow is especially well utilized by spiking neural networks (SNNs) - computational abstractions of biological neurons that transmit signals in an all-or-none spiking fashion only when the internal neuronal activation reaches a certain threshold \cite{Maass_97}(Figure \ref{fig:Figure1}). Indeed, asynchronous all-or-none spikes seem like a perfect match to asynchronous all-or-none events due to the similar nature of these signals. Studies showed that when it comes to optical flow SNNs can perform on par \cite{hagenaars2021selfsupervised} or even better \cite{Kosta23,Cuadrado} than state-of-the-art traditional neural networks, while requiring up to 50-times fewer parameters \cite{Kosta23} and promising up to $10^3$ decrease in compute energy \cite{Zhang23}. With the development of specially designated to processing SNNs neuromorphic processors \cite{Davies_etal18}, event-based optical flow estimation with SNNs reached a significant milestone of technological maturity - implementation for vision-based navigation onboard a drone in a fully neuromorphic pipeline \cite{paredesvallés2023fully}.

Although theoretically, SNNs are capable of performing a wide variety of tasks \cite{mcculloch1943logical,HORNIK1989359,MAASS19971659,Maas_fast}, their actual performance strongly depends on the choice of proper priors such as neural network architecture and connectivity on the hardware and algorithmic levels. For example, in image classification tasks, the introduction of randomly initialized and non-trainable convolutional kernels improves the classification accuracy compared to fully connected networks \cite{Frenkel2021} simply because convolutional kernels extract a local spatial pattern. Similarly, proper priors to extract spatio-temporal patterns could enhance algorithms of the optical flow estimation from event-based data. 

A good computational prior for optical flow estimation from event-based data could be the correlation-based motion detector, which is ubiquitously employed in the animal kingdom \cite{clark2016parallel}. As its name suggests, this detector extracts local motion direction and velocity (i.e. optical flow) based on the temporal correlations between different points in space \cite{clark2016parallel}. Such correlation-based estimates were shown to closely approach the optimal way to extract motion signals in conditions with low Signal-to-Noise Ratio (SNR)\cite{Sinha,Potters_1994}. Event-based cameras, meanwhile, are prone to noisy outputs \cite{guo2023} due to the environment (low contrast and light intensity \cite{Brosch2015}), the temperature \cite{Czech2016,Nozaki}, the junction leakage \cite{Nozaki,Ding_2023}, the hardware limitations (manufacturing mismatch \cite{Lichtsteiner2008}), and the very differential nature of processing which amplifies high-frequency noise \cite{Ding_2023,Brosch2015,Simoncelli1993DistributedRA}. Given such low SNR output, a correlation-based motion detector is well-suited to process event-based data.

Bio-inspired correlation-based methods for optical flow estimation with SNNs generated long-lasting interest in event-based vision and neuromorphic computing communities with over a decade of relevant research on both software \cite{Paredes_Valles_2020,barranco2015bio, Brosch2015,zheng2023spike,orchard2013spiking} and hardware levels \cite{brosch2016event, Milde, Schoepe2024,giulioni2016event,dangelo2020event,gutierrez-galan2022event,haessig2018spiking,Chiavazza_etal23,Sarpeshkar,Kramer,EtienneCummings}.  In particular, for our study, we are interested in the inspired by the classical model of the insect motion detector  \cite{reichardt1961autocorrelation}  Time-Difference Encoder (TDE) \cite{Milde} due to its circuit and algorithmic simplicity \cite{Milde}. The TDE detects motion by comparing signals at two points in space: input to the facilitator provides a time-decaying gain, whereas input to the trigger converts gain to an input current impulse to a Current-Based Leaky Integrate-and-Fire (CuBa LIF) neuron, which integrates it in its membrane potential and converts it into a firing activity as shown in Figure \ref{fig:Figure1}. The TDE tuning to the preferred direction of motion (PD) is achieved since it produces output spikes only when the trigger event arrives after the facilitator one, as otherwise there is no gain to be converted to firing activity. The TDE was successfully implemented on FPGA \cite{gutierrez-galan2022event}, digital neuromorphic hardware such as SpiNNaker \cite{dangelo2020event}, Loihi \footnote{TDE implementation on Intel Loihi: https://github.com/intel-nrc-ecosystem/models} and Loihi 2 \cite{Chiavazza_etal23}, and custom mixed-signal neuromorphic hardware \cite{Milde}. Its potential was demonstrated in robotics applications for collision avoidance \cite{Milde,Schoepe2024} and estimation of ego-motion \cite{Chicca25}. The strength of the underlying computational primitive was also explored in tasks that are not related to vision, such as spotting keywords \cite{nilsson2023comparison}.

Despite its appeal, the classical two-point TDE (TDE-2) has a notable deficiency: in highly textured environments, it loses direction selectivity because motion in a non-preferred direction provides the TDE with residual activity in the gain compartment, which can be subsequently triggered by another edge moving in the non-preferred direction as shown in Figure \ref{fig:Figure1}. Given that successful event-based navigation requires events that are produced by the apparent motion of textures\cite{paredesvallés2023fully}, loss of direction-selectivity makes classical TDE-2 unsuitable for vision-based navigation tasks that require precision in optical flow estimation. Of course, this loss in direction-selectivity can be partially compensated by either max pooling \cite{Chiavazza_etal23}, Winner-Take-All mechanism \cite{Milde,Schoepe2024} or subtraction of the velocities estimated by opposingly tuned detectors. Yet in that case, the energy-efficiency of the circuitry would be obstructed as individual TDEs would still respond to the non-preferred motion.

To solve this problem, we propose to follow recent findings in neuroscience\cite{Haag}  and to augment the TDE with a third, inhibitory input. While the classical model of elementary motion sensitivity in insects \cite{reichardt1961autocorrelation,clark2016parallel} estimated temporal correlations from two spatial locations, Haag et al. \cite{Haag} showed that insect elementary motion detectors (so-called T4 and T5 cells) pools signals from three spatial locations with one of them being inhibitory and activated by the motion in ND. In our implementation, activation of this inhibitory input resets the gain to zero and eliminates the residual activity to retain the TDE direction-selectivity (Figure \ref{fig:Figure1}). Consequently, even in highly textured environments, individual TDEs retain direction selectivity, and the entire circuit spikes much less, becoming more energy-efficient. We propose to call this detector TDE-3. 

Upon fixing issues related to detector architecture, we performed a more in-depth study of the TDE coding of velocity with synthetic and real-world data.  Specifically, there are five major contributions of this paper:

\begin{enumerate}
    \item We present a  more robust TDE-3 and highlight its advantage over the TDE-2 on the level of direction-selectivity of a single detector (Figures \ref{fig:Figure1} \& \ref{fig:Figure6}).
    \item We develop a pipeline for the supervised training of the TDE using Back-Propagation Through Time (BPTT) and surrogate gradients \cite{Neftci_etal19, zenke2021remarkable} to linearly map input velocities into an output spike count (Figure \ref{fig:Figure2}) or an inter-spike interval (ISI) between the first and the second spike (Figure \ref{fig:Figure3}) within both wide (10-fold, Figure \ref{fig:Figure7}) and narrow (1.5-fold, Figure \ref{fig:Figure8}) dynamic range with precision of up to 2\% (Table \ref{tab:narrow}). 
    \item We develop a novel approach for training SNNs to have a specific ISI using BPTT. While most other methods \cite{BOHTE200217, Hesham, Comsa} model the relationship between input magnitude with differentiable function than use BPTT, we measure ISI from the amplitude of low-pass filtered spike train to use BPTT with surrogate gradient (see Section \ref{subsec:isi} for the details). The theoretical advantage of our method is that it does not require estimation of the relationship between spike timing and input magnitude, allowing work with more conventional activation functions. 
    \item We compare the robustness of velocity coding with TDEs to noise (Figures \ref{fig:Figure10}, \ref{fig:FigureNew}) and spatial frequency (Figure \ref{fig:Figure9}) composition of the scene - two major challenges for velocity coding with correlation-based motion detectors. Our results show that spike count is more robust to noise, while ISI is less affected by changes in spatial frequency and provides faster velocity inference. 
    \item We performed the first-ever quantitative characterization of resolution, robustness, and energy efficiency of encoding real-world optical flow by TDEs (Figures \ref{fig:Figure11}, Figures \ref{fig:Figure12} and Tables \ref{tab:boxes}, \ref{tab:disk}). Indeed, although Schoepe et al. \cite{Schoepe2024} quantitatively assessed the dependence of the TDE output on the velocity of real-world moving gratings, they did so on the network level and by measuring activity over the entire sequence rather than based on individual detector activations.
    Our results show that the TDE performs on par with model-based methods of local motion estimation reported in the literature (\cite{rueckauer2016,Gallego_etal22} ( directional error < 20\textdegree), but has the advantage of being naturally compatible with spike-based neuromorphic hardware. Furthermore, the TDE-3 is more energy-efficient than the TDE-2, requiring up to $2-4 \times$ fewer spikes (Table \ref{tab:boxes}, Supplementary Information). 
\end{enumerate}
 
To sum up, the paper presents a robust computational primitive to optimize the estimation of optical flow from event-based data, compatible with spike-based neuromorphic hardware for low-power inference.

\section{\textbf{Time-Difference Encoders: Principles of Operation}} \label{subsec:TDE}

Before discussing the main findings of the paper it is important to understand the TDE-2 and TDE-3 computational principles on the level of direction-selectivity and velocity coding to get insight into the features and limitations of the systems under investigation.

\subsection{\textbf{Computational primitive and detector architecture}}
The TDE-2 computational principle is to extract temporal correlations at two spatial locations: facilitator and trigger. A spike input to the facilitator provides the detector with a time-decaying gain, while a spike input to the trigger converts the gain into an excitatory current which charges CuBa-LIF neuron and leads to spiking \cite{Milde} (Figure \ref{fig:Figure1}). Essentially, the time-decaying gain extracts temporal correlations as the neuron reports the coincidence of inputs to the facilitator and trigger within a temporal window of the gain decay and scales the current according to the delay between activation of the facilitator and trigger. Hence, faster motion leads to a higher number of spikes with shorter intervals between them and vice versa. 

The direction-selectivity of the neuron is ensured by the combination of the two factors. Its activation in the direction anti-parallel to PD, also known as null direction (ND), is prevented since when activation of the trigger precedes the activation of the facilitator, there is no gain to convert to current (Figure \ref{fig:Figure1}A center). In the case of the motion in the direction orthogonal to PD (OD), the facilitator and trigger are activated simultaneously. In the hardware, detector firing is prevented by the delay between the processing of the trigger and gain. Here the delay is realized through the sequence of computational steps: TDE-2 first processes trigger and then gain. Hence, in the case of OD motion, there is also no gain to be converted to the current at the moment of trigger activation.

The crucial deficiency of such a system is that upon motion in OD or ND, there is no gain to be converted to current \textbf{only} at the moment of the trigger's activation by the same stimulus. As in such purely excitatory architecture, motion in OD and ND leaves a trace of residual gain activity, another stimulus moving in a non-preferred direction will lead to detector activation (Figure \ref{fig:Figure1}A). Hence, upon motion in a highly textured environment, individual TDE-2 can lose direction-selectivity. 
 
Our TDE-3 solves the problem of the residual gain by incorporating a third, inhibitory input. Relative to the gain, the inhibitory input is located on the opposite side of the detector and upon activation resets the gain to zero (Figure \ref{fig:Figure1}). As a result, when secondary stimuli are moving in the ND, they will first appear at the inhibitory flank. This will remove any residual gain activity, such that when the stimulus activates the trigger, there again will be no gain to be converted to current. In the case of motion in the OD, detector activation will be prevented by simultaneous processing of facilitator and inhibitory inputs (assuming stimuli wide enough to cover 3 inputs) such that the OD motion will leave no trace in gain activity. Thus, the addition of the inhibitory flank allows the detector to retain direction-selectivity in textured environments (Figure \ref{fig:Figure6}). As a consequence of such more robust direction-selectivity of individual TDEs, the entire network can spike up to 2-4 times less (Table \ref{tab:boxes}, Supplementary Information Figure S2).

\subsection{\textbf{Velocity coding with TDE}}

When the moving stimulus passes the TDE, two properties of the emitted spike train allow downstream circuitry to estimate velocity: the number of spikes and ISI \cite{Milde, gutierrez-galan2022event}.  The spike count is directly proportional to the input velocity, while the ISI is inversely proportional. Indeed, fast motion means a short temporal delay between the activation of the gain and its conversion to current. Consequently, the current is large and the number of emitted spikes is high. Conversely, such vigorous firing means that spikes closely follow each other and the interval between them is small \cite{Milde}. Of special interest is ISI between the first two spikes as it allows very fast velocity inference \cite{Milde}.

To decode stimulus velocity from TDE activity in the presence of multiple edges one needs to perform motion segmentation to relate counted spikes/measured ISIs to a particular moving edge. To do so we propose to count spikes starting from each activation of TDE's trigger as it indicates passage of a new edge. Additionally, we limited the time window over which spikes are counted for spike count inference. Otherwise, in the presence of multiple edges motion of the very last edge would affect the measured velocity of the very first edge.

From the perspective of a single neuron's coding capacity, and regardless of the inference method, all-or-none spiking is more of a liability than an asset. This is because it introduces a latency vs. dynamic range dilemma, as spiking occurs in time  \cite{Rieke99}. For example, to discriminate input velocities within a range of, e.g., 100 different values would require a coding window of at least 100 timesteps either to count spikes or to measure interspike intervals. Even with a very short timestep of 1 ms, the lag becomes substantial for edge applications like robotics, where neuromorphic hardware is needed the most and low latency is paramount. Thus, to avoid impractical latency, one can either encode a narrow range of input velocities with high resolution or a wide range of input velocities with low resolution (Figures \ref{fig:Figure7}, \ref{fig:Figure8}). Therefore, a strategy to reconcile latency and dynamic range requirements would be to use each detector within a relatively narrow range of velocities, while boosting the dynamic range of the entire network by using detectors with different spacing between detector inputs.

The coding of velocity by the TDE is additionally troubled by the noise and spatial frequency of a scene. Noise leads to the non-stimulus-related activation of detectors. Spatial frequency bias detector responses as edges closely following each other will lead to stronger TDE activation than a single edge. In the results section, we compare spike count and ISI-based inference in terms of their robustness in various noise (Figures \ref{fig:Figure10}, \ref{fig:FigureNew}) and spatial frequency (Figure \ref{fig:Figure9}) conditions.

\section{\textbf{Experiments and results}} \label{sec:experiments}
The experiments that we performed can be classified into three categories: (1) investigation of the robustness of the direction-selectivity of individual TDE-3 and TDE-2 (Figure \ref{fig:Figure6}), (2) TDEs training to linearly map input velocities into output (Figures \ref{fig:Figure7}-\ref{fig:Figure10}), and (3) application of the TDEs for optical flow inference from real-world data (Figures \ref{fig:Figure11}, \ref{fig:Figure12}).

\subsection{\textbf{Robust direction-selectivity of individual TDE-3 in textured environments}}
\begin{figure}[h]
\centering
\includegraphics[width=0.8\textwidth]{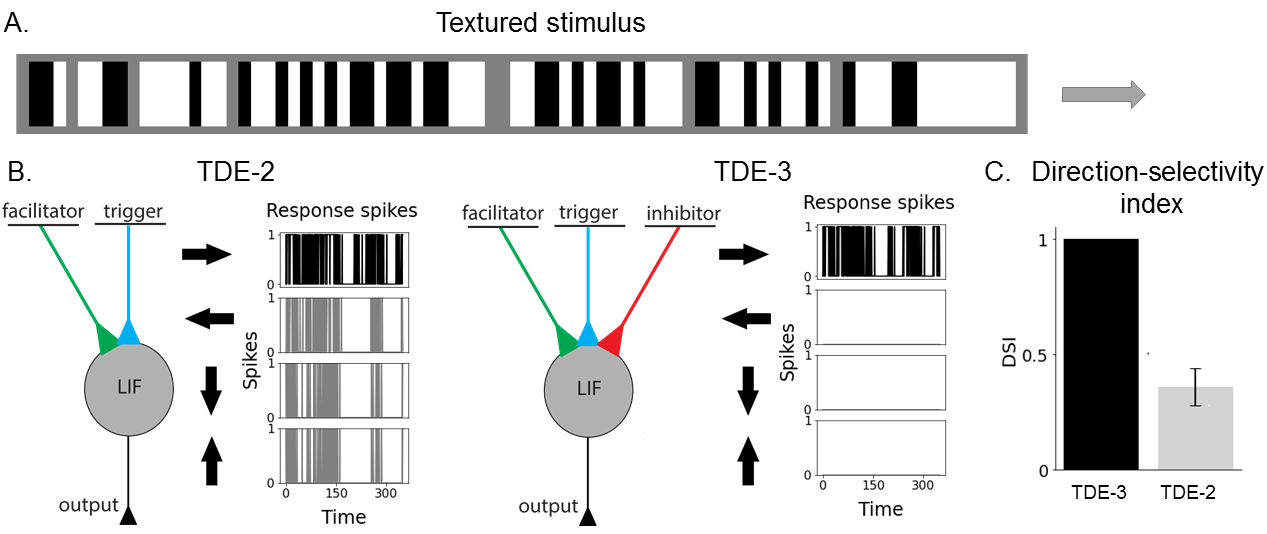}
 \caption{\raggedright Augmented 3-point TDE retains direction-selectivity in a textured environment. A - \textit{Visual stimulus composed of vertical bars. The bars had three light intensity levels: white, grey, and black. The size of the texture along the motion axis was 80 pixels and 3 pixels along the orthogonal direction. We employed 5 velocities: 0.1 px/timestep, 0.2 px/timestep, 0.33 px/timestep, 0.5 px/timestep and 1 px/timestep. The motion direction (left-right, right-left, top-bottom, bottom-top) and velocity were randomly chosen for each stimulus example (2000 examples per testing round, 400 testing rounds). To vary the "amount" of texture in stimuli we randomly varied the fraction of the grey bars from 0\% to 80\%.} B. TDE-2 and TDE-3 and their responses to stimuli moving in 4 cardinal directions. C. Direction-selectivity index (fraction of spikes fired upon stimulus motion in PD.}
\label{fig:Figure6}
\end{figure}

To compare the direction-selectivity of TDE-2 and TDE-3, we simulated their responses to textures moving in four cardinal directions (Figure \ref{fig:Figure6}B): left-to-right (L-R), right-to-left (R-L), top-to-bottom (T-B) and bottom-to-top (B-T). For each stimulus presentation (2000 per testing round), we randomly varied the texture composition, motion direction, and velocity (see Section \ref{subsec:data} for the details). 
To show that the robust direction-selectivity of the TDE-3 stems from its structure and is not due to some particular set of parameters, we varied at each testing round (400 in total) all of the TDE parameters over a 10-fold range. Without loss of the generalization, we limited our investigation of direction-selectivity to responses of  L-R motion-sensitive TDEs (Figure \ref{fig:Figure6}B) that sampled signals from three adjacent pixels through which textures were moving.

Figure \ref{fig:Figure6}B shows how the responses of TDE-2s and TDE-3s depend on the texture motion direction. One can appreciate that while L-R TDE-2 fires to all stimuli directions, the spiking of the TDE-3 is contained to stimuli in its preferred direction. 
To assess the TDE's direction-selectivity quantitatively, we calculated the so-called Direction-Selectivity Index (DSI) as the ratio between the number of spikes emitted by motion in PD and the total number of spikes fired by the TDE in a testing round.
Figure \ref{fig:Figure6}C compares the DSI of TDE-2s and TDE-3s averaged across all of the testing rounds. 
For the TDE-2, the mean DSI is 0.36 with a standard deviation of 0.08. This indicates a rather poor direction-selectivity in a textured environment, as on average inference of motion direction with such a detector is only 11\% better than random guessing. On the other hand, the TDE-3 has a DSI of exactly 1, with zero deviation. Indicating, that regardless of the initialization of the parameters, simply because of its structure with inhibitory input that eliminates residual gain activity (details - Section \ref{subsec:TDE}, illustration - Figure \ref{fig:Figure1}), the TDE-3 is activated only by motion in the PD. 

For simplicity of interpretation, in Figure \ref{fig:Figure6} we limited our analysis to 1D motion. Yet, the same observation is true for the 2D motion (Supplementary Figure S1). We performed experiments with textures of increasing complexity (vertical bars, checkerboard, randomly generated textures) moving in 2D and showed that in all of the cases, TDE-3 had higher DSI than TDE-2. However, an increase in texture complexity decreased the DSI of TDE-3 (see Supplementary Figure S1 for the details).

\subsection{\textbf{Supervised training of TDE-3}}
To promote linear mapping between the input velocities in PD and the TDE outputs, we performed supervised training with BPTT and surrogate gradients for both spike count and ISI-based inference methods and compared obtained results (see Section \ref{subsec:TDE} and \ref{subsec:training} for the details). Note that in the context of linear mapping of input velocities into TDE outputs, the TDE-2 and TDE-3 are identical because the delay between the activation of the facilitator and trigger is not affected by the inhibitory input. Therefore, although in experiments with synthetic data we perform all of the training on the TDE-3, the results are generalizable to the TDE-2. 

There were three types of experiments with training of TDE-3.
First, we established that detectors are trainable in principle and that linear mapping between input velocities in a given cardinal direction and TDE output is possible for various stimuli dynamic ranges (Figures \ref{fig:Figure7} and \ref{fig:Figure8}, Table \ref{tab:narrow}).
Second, we investigated how training and mapping are affected by changes in the spatial frequency of the stimulus - a known liability of correlation-based motion detectors \cite{BORST2011974} (Figure \ref{fig:Figure9}). 
Third, we studied the robustness of TDE training and inference in the presence of various levels of noise (Figures \ref{fig:Figure10}, \ref{fig:FigureNew})

\subsubsection{\textbf{Linear mapping of velocity of moving edge}}
To study how the TDE can map the input velocity into its output, we simulated the TDE responses to a single edge moving from left to right at various constant velocities. We opted for these stimuli because they directly target the goal of linear mapping while keeping things simple and unambiguous. To avoid overfitting and the necessity to separate stimuli into validation and test datasets, we randomly assigned the velocity of an edge at each stimulus example presentation. 
We employed two stimuli sets. One contained 5 velocities within a 10-fold range of magnitude and was used to show that the TDE can encode velocities in a wide dynamic range. The second stimuli set contained 15 different velocities spanning ~1.5-fold range of magnitudes and was used to show that the TDE can encode velocities with fine resolution (see Section \ref{subsec:data}). 

As a metric of the quality of linear mapping, we used the Pearson correlation coefficient between the stimulus and the estimated velocity \cite{Pearson1895}. The performance of the trained TDEs was assessed with both ISI and spike count regardless of which one was used during training to study the connection between these two inference methods.  Additionally, to assess the dynamic energy efficiency of the trained detectors, we calculated the average number of spikes elicited by a stimulus. 

\begin{figure*}[h]
  \centering
  \includegraphics[width=1.0\linewidth]{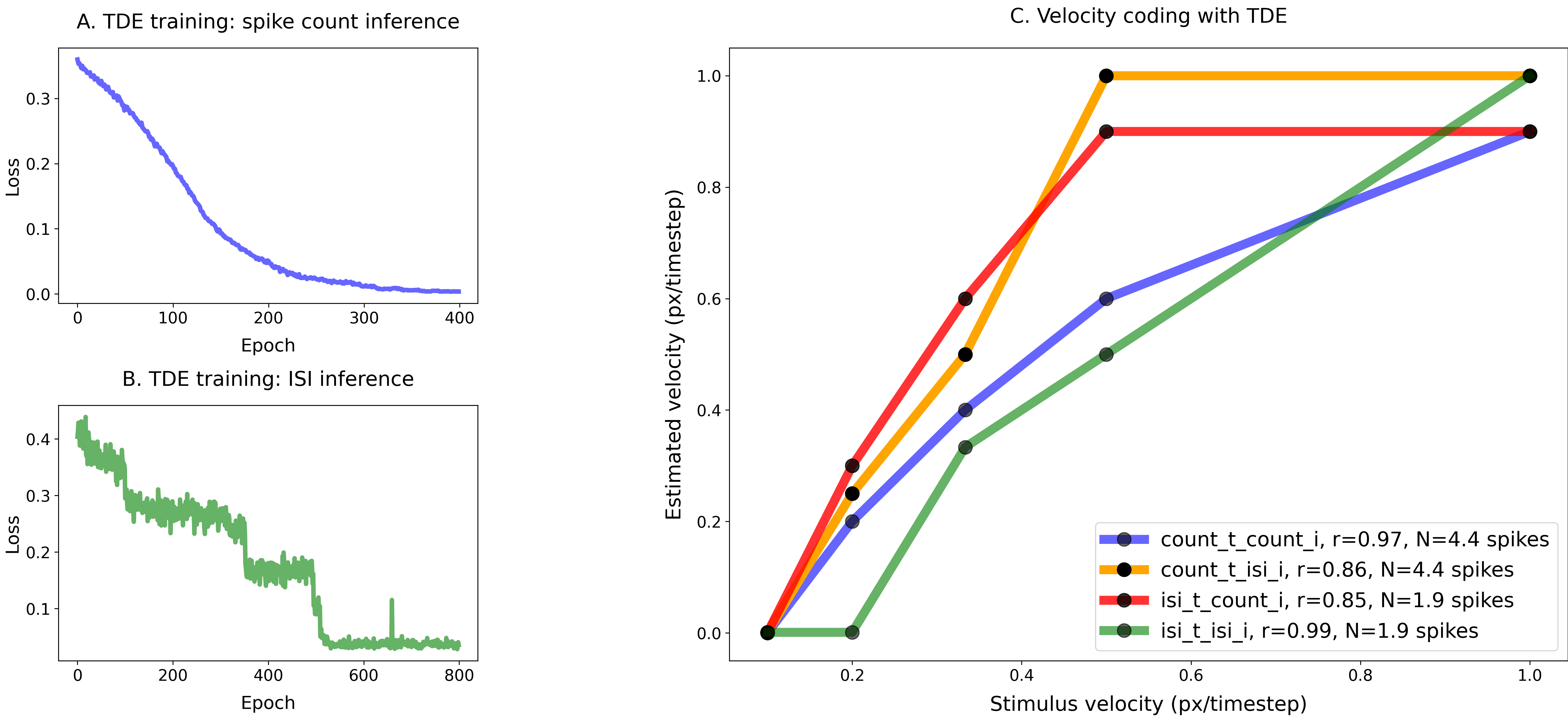}
  \caption{Training of the TDE-3:Wide dynamic range, low resolution, one moving edge (A)Loss function for spike count-based inference, (B) loss function for ISI-based inference, (C) comparison of the velocity tuning curves. Blue - training with spike count, inference with spike count during the test; Orange - training with spike count, inference with ISI during the test; Green - training with ISI, inference with ISI during the test; Red - training with ISI, inference with spike count during the test.}
  \label{fig:Figure7}
\end{figure*}

Figure \ref{fig:Figure7} shows that the TDE can be trained to encode a wide dynamic range of input velocities. The loss function on panels A (training with the spike count) and B (training with ISI) gradually decreases during training, indicating that with both inference methods it is indeed possible to train the TDE to discriminate input velocities in a wide dynamic range. 

Panel Figure \ref{fig:Figure7}C makes a quantitative comparison of velocity coding with various training and inference methods. Unsurprisingly, the correlation coefficient between estimated and true velocity is the highest when evaluation is done with the same inference method as training: r=0.97 for the spike count training and inference (blue line), r=0.99 for the ISI training and inference (green line). Yet training with spike count leads to reasonable performance with ISI inference (r=0.86, orange line) and vice versa (r=0.85, red line) with the only problem being distinguishing between the two highest velocities. Training with ISI-based inference also leads to a lower average number of spikes emitted by stimulus (1.9 vs 4.4), suggesting higher energy efficiency when implemented on neuromorphic hardware \cite{caccavella_etal23}.

\begin{figure*}[h]
  \centering
  \includegraphics[width=1.0\linewidth]{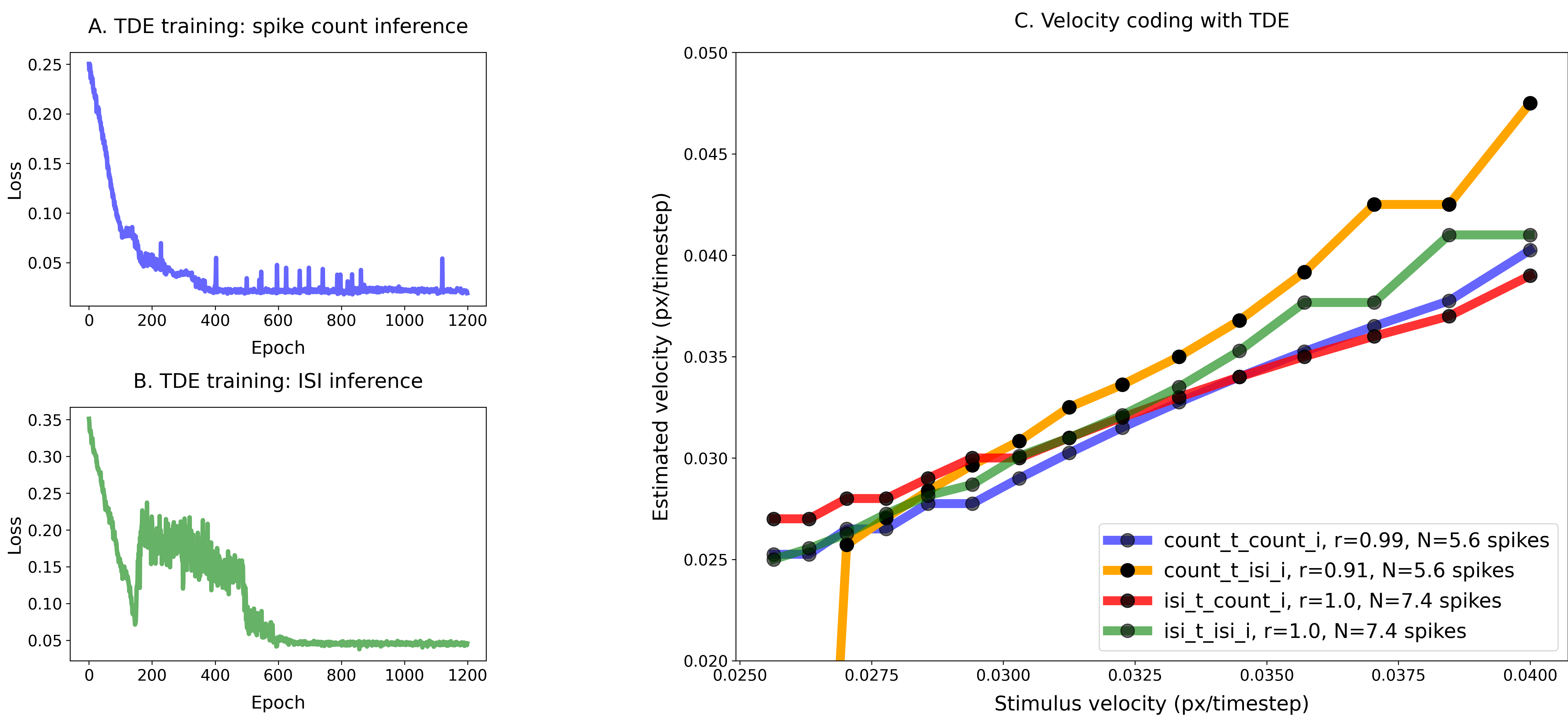}
  \caption{Training of the TDE-3: Narrow dynamic range, high resolution, one moving edge. Training of the TDE-3: wide dynamic range, low resolution. (A)Loss function for spike count-based inference, (B) loss function for ISI-based inference, and (C) comparison of the velocity tuning curves. Blue - training with spike count, inference with spike count during the test; Orange - training with spike count, inference with ISI during the test; Green - training with ISI, inference with ISI during the test; Red - training with ISI, inference with spike count during the test.}
  \label{fig:Figure8}
\end{figure*}

Figure \ref{fig:Figure8} demonstrates that the TDE can effectively encode a narrow dynamic range of input velocities with high precision (Panels A and B). In Panel C, one can appreciate that apart from the case of training with spike count and evaluation with ISI (orange line, \(r = 0.91\)), input velocities are mapped into TDE outputs nearly perfectly, with a correlation \(r \approx 1.0\) (Table \ref{tab:narrow}). Also, ISI evaluation with spike count training results in a decrease in dynamic range, as the estimate becomes zero for the first two velocities. This occurs because these two almost equal velocities (0.0256 and 0.0263) were encoded with 1 spike, making the ISI indeterminable.

For this experiment, fine resolution was paramount. Therefore, we calculated mean relative error by dividing the difference between true and inferred velocity by true velocity and averaging this metric across all velocities (see Equation \ref{eq:equation7}, Table \ref{tab:narrow}). 
Unsurprisingly, ISI evaluation with spike count training yielded the largest relative error (30\%). Even when excluding the first two velocities, which because TDE emitted only 1 spike contributed to the variance the most, the relative error remains >= 3 times higher (7.9\%) than for the other three conditions, where velocity was encoded with an error as small as 2.1\% (ISI training, spike count evaluation, red line). 
We also note in this experiment that training with ISI led to a higher average number of spikes (7.4) compared to spike count training (5.6) (see Table \ref{tab:narrow}).

To conclude, the TDE can be trained to successfully encode the velocities in wide and narrow ranges.

\begin{table}[H]
   \caption{High-resolution velocity inference with the TDE.}  
   \label{tab:narrow}
   \small
   \centering
   \begin{tabular}{lcccr}
   \toprule\toprule
   \textbf{Training Inference} & \textbf{Test Inference} & \textbf{Correlation, \textnormal{\textbf{r}}} & \textbf{Relative Error (\%)} & \textbf{N Spikes} \\ 
   \midrule
   \multirow{2}{*}{spike count} & spike count & 0.99 & 2.5 & \multirow{2}{*}{5.6} \\
                                  & ISI & 0.91 & 30(7.9\textsuperscript{\hyperref[asterisk]{*}})& \\
   \midrule
   \multirow{2}{*}{ISI} & spike count & 1.0 & 2.1 & \multirow{2}{*}{7.4} \\
                         & ISI & 1.0 & 2.7 & \\
   \bottomrule
   \end{tabular}
\end{table}
\small\textsuperscript{*}\label{asterisk} Here first two input velocities generated only 1 spike. Therefore the estimated velocity was 0, which led to a large error in the estimated velocity. We excluded those two values from the analysis to get a better fill of what is actual coding error when the velocity can be estimated and get the value of 7.9\%

\subsubsection{\textbf{Spatial frequency and velocity coding by TDE}}
When multiple moving edges are present, temporal signal integration by current and voltage compartments could lead to biases in velocity estimation due to variations in distance between moving edges (i.e. spatial frequency). 
To test whether training of the TDE-3 can mitigate this problem, we trained it to linearly map the velocities of two moving edges into the spike count and ISI while varying the distance between edges in a 10-fold range (see Section \ref{subsec:data}). Here we employed a wide dynamic range of velocities (see Section \ref{subsec:data}) as it presents a more natural kind of stimuli since optical flow in real-world scenes is known to vary over a large range of apparent velocities \cite{mitrokhin2019evimo,rueckauer2016}.  

\begin{figure*}[h]
  \centering
  \includegraphics[width=1.0\linewidth]{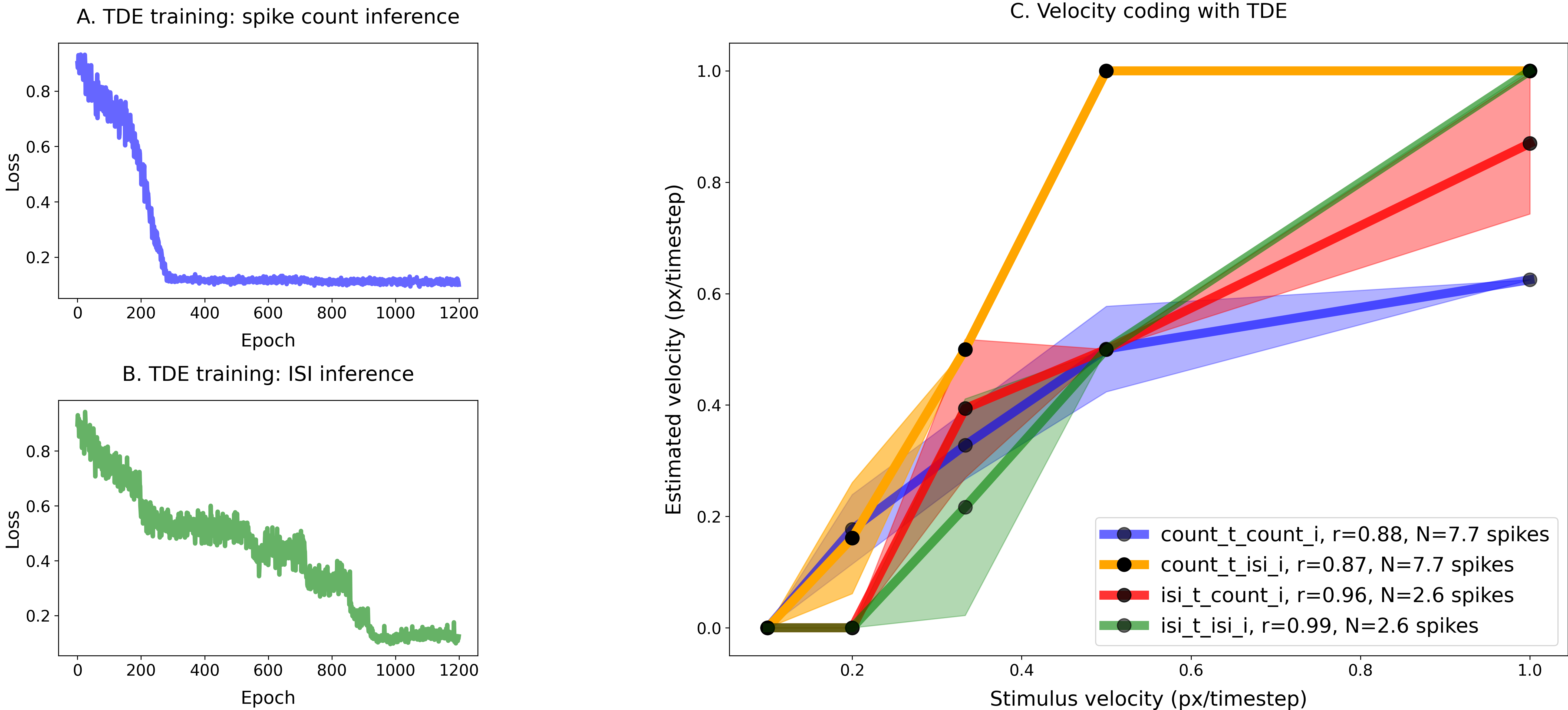}
  \caption{Training of the TDE-3: Robustness to variation in spatial frequency. Wide dynamic range, low resolution, two moving edges at randomly assigned distances.  (A)Loss function for spike count-based inference, (B) Loss function for ISI-based inference. (C)Comparison of the velocity tuning curves. Blue - training with spike count, inference with spike count during the test; Orange - training with spike count, inference with ISI during the test; Green - training with ISI, inference with ISI during the test; Red - training with ISI, inference with spike count during the test. As there was no noise, all of the variation in TDE response (shading) originates from the variation in stimulus spatial frequency (spacing between the edges)}
  \label{fig:Figure9}
\end{figure*}

Figure \ref{fig:Figure9} shows that one can indeed train the TDE to linearly encode velocities of multiple moving edges with both spike count- (Panel A) and ISI-based (Panel B) inference. In Panel C, we plotted the obtained tuning curves of the responses to the second edge for all 4 testing combinations. Shading indicates standard deviation in response to a given velocity and originates from the variation in distance between edges. Panel C shows that regardless of inference during testing, training with ISI (ISI test -green line, r=0.99; spike count test -red line, r=0.96) leads to a higher correlation with stimulus velocity than training with spike count-based inference (spike count test - blue line, r=0.88; ISI test - orange line, r=0.87).
However, training with ISI-based inference tends to constrict dynamic range, effectively encoding velocities over a 3-fold range. Training with spike count-based inference preserved almost the entire dynamic range of input velocities but saturated at the highest velocities. Training with spike count also led to a higher average number of emitted spikes 7.7 vs 2.6. Noteworthy, regardless of the inference method employed during training, spike count-based inference during testing resulted in higher variation in the estimation of velocity. 

To conclude, Figure \ref{fig:Figure9} shows that the trained TDE is fairly robust to the variation in stimulus spatial frequency. 

\subsubsection{\textbf{Noise and velocity coding by TDE}}

\begin{figure*}[h]
  \centering
  \includegraphics[width=1.0\linewidth]{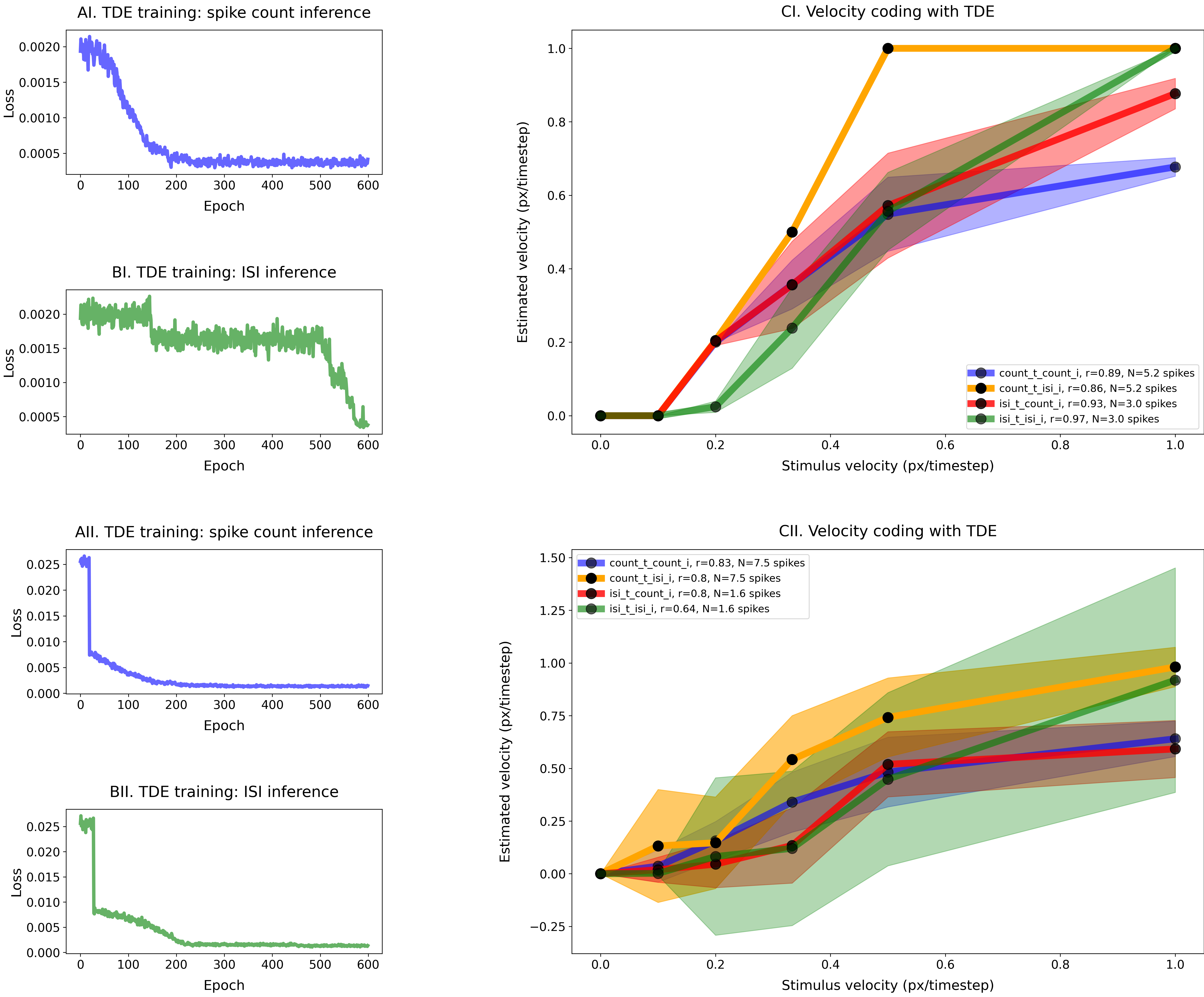}
  \caption{Training TDE: Performance in the presence of noise. Top row - low noise, Bottom row - high noise. (A)Loss function for spike count-based inference, (B) loss function for ISI-based inference, and (C) comparison of the velocity tuning curves. Blue - training with spike count, inference with spike count during the test; Orange - training with spike count, inference with ISI during the test; Green - training with ISI, inference with ISI during the test; Red - training with ISI, inference with spike count during the test.}
  \label{fig:Figure10}
\end{figure*}

To test whether the training of the TDE to encode velocity is robust towards noise, we injected into our simulation background activity noise - one of the primary sources of noise in event-based vision \cite{guo2023}. The noise was injected as stochastic events with mean rates varied 50-fold in magnitude: from 0.2 Hz/px to 10Hz/px (see Section \ref{subsec:data}). 

 To allow the network to combat noise we employed a Spatio-Temporal Correlation Filter (STCF) of events (Section \ref{sec:Methods}). STCF is a popular noise-combating tool in the field of event-based vision \cite{guo2023,rueckauer2016,Gallego_etal22}  that allows an event to pass the filtering stage only when there are events in at least n neighboring pixels within a specified timeframe (Section \ref{subsec:training}, \cite{guo2023}) with n being a trainable parameter. Although as a result we would, strictly speaking, test not the performance of the TDE, but of the TDE and STCF, our approach is still sound. The reason for this is that STCF is almost always part of the TDE circuitry anyhow \cite{Milde,Schoepe2024,dangelo2020event,gutierrez-galan2022event,Chiavazza_etal23}. Therefore, since in real applications TDE will be always paired with STCF, it makes sense to investigate their joint performance. Especially, given that STCF does not eliminate all of the noise and there is always a concern that aggressive filtering would destroy part of the signal. 
 
In the presence of noise, not every increase in current is caused by a moving edge, therefore one of the goals of the training is to suppress the TDE activations caused by noise. To account for this factor in the calculation of the loss, we calculated the loss between all of the velocity estimates performed by the TDE and stimulus ground truth at corresponding timepoints (effectively, this ground truth was zero everywhere except at the moment of edge appearance). Also, to evaluate how the trained TDE can reject noise and preserve true motion signal, we employed an additional metric called fraction of true activity (FTA) as the sum of estimated velocities caused by the stimulus (true motion signal) to the total sum of estimated velocities (caused by signal and noise). 
 
\begin{figure*}[h]
  \centering
  \includegraphics[width=1.0\linewidth]{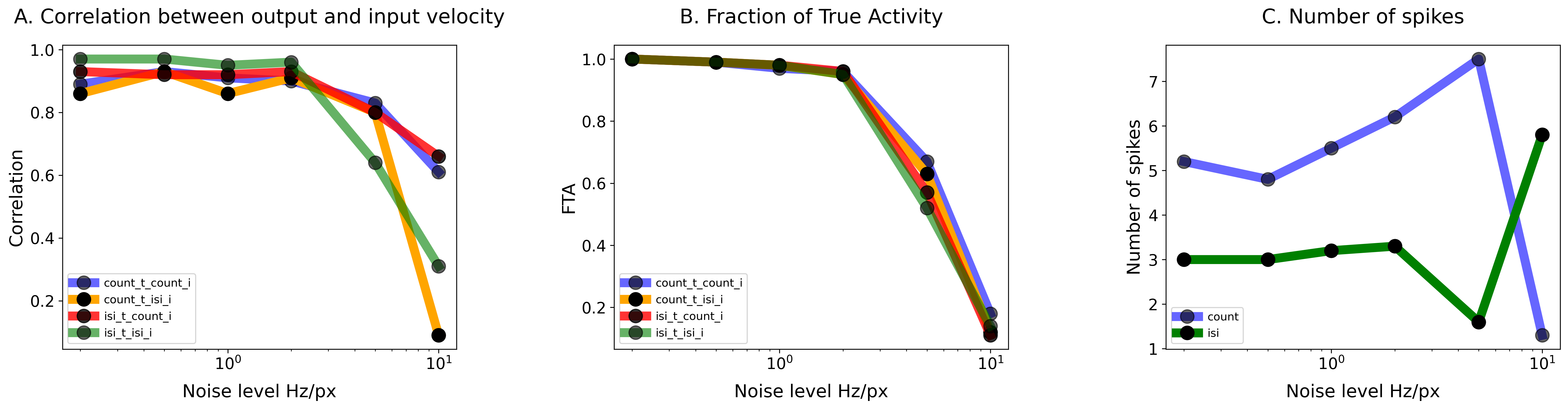}
  \caption{Training of the TDE-3: Summary statistics of TDE performance in the presence of noise. A: Correlation of the TDE output with ground truth input velocity as a function of noise magnitude. B: FTA as a function of noise magnitude. C: Average number of spikes used for encoding of an edge velocity as a function of noise level. In all three cases, noise levels are plotted on the logarithmic scales. For panels A and B the color code is as follows: Blue - training with spike count, inference with spike count during the test; Orange - training with spike count, inference with ISI during the test; Green - training with ISI, inference with ISI during the test; Red - training with ISI, inference with spike count during the test. For panel C blue describes training with spike count, green - training with ISI.}
\label{fig:FigureNew}
\end{figure*}

Figure \ref{fig:Figure10} shows the results of the training for two extreme cases of "bright" (noise rate=0.2Hz/px) and "dark" (noise rate=5Hz/px) scenes. With the low noise level, the performance of the TDE (Figure \ref{fig:Figure10} top row) was almost identical to the performance in noiseless experiments (Figure \ref{fig:Figure9}) in terms of shapes of tuning curves and correlation with stimulus velocities. These observations are also confirmed by FTA (Figure \ref{fig:FigureNew}) with nearly all TDEs activity being caused by the stimulus. 
Results of the training with a high level of simulated noise are shown in Figure \ref{fig:Figure10}, bottom row.  With both inference methods, there is an early dip in the loss function (Panels AII and BII) caused by an increase in the number of neighboring pixels where at least one event ought to occur for any given event to pass the STCF processing stage (see Section \ref{subsec:training}).  As a result of this, more noise is rejected and loss dramatically decreases. 

Panel CII illustrates tuning curves obtained in all 4 testing conditions. One can immediately note that training and inference with ISI (green line) leads to massive standard deviation (shading) in the estimated velocity. 
Surprisingly, TDEs trained with ISI showed better performance when testing was done with spike count (red line) as the standard deviation was lower and the correlation coefficient was higher (r=0.8 vs r=0.64). 
With ISI-based training, about half of the encoded motion signals were related to the stimulus (Figure \ref{fig:FigureNew}). Yet, given the low mean spike count (N=1.6), TDEs coding capabilities were rather poor with at least one edge being ignored most of the time. Training with spike count-based inference leads to much better performance both with spike-count (blue line and shading, r=0.83) and ISI-based (orange line and shading, r=0.8) evaluation. 
Spike count-based training leads to a mean spike count of 7.5 spikes. Since with this training condition almost 70\% of encoded motion signals were related to the stimulus (Figure \ref{fig:FigureNew}), we conclude that spike count-based training allows TDE to fairly reliably encode motion signals in the presence of noise. 

In Figure \ref{fig:FigureNew} we summarized the influence of the noise levels on the correlation between output and ground truth input velocity (panel A),  FTA (panel B), and mean number of spikes generated by a moving edge (panel C). Panels A and B show that TDE-3 is very robust up to a noise level of up to 2Hz/px with  FTA >95\% and correlation with the stimulus velocity r of $\approx\ 90\%$ throughout and largely regardless of the training / inference method. However, a further increase in noise level led to a sharp decrease in performance with most of the TDE output being noise-driven. Our results also show that in the presence of noise, the spike-count-based inference serves as a more robust and reliable training signal. The likely reason for this is that ISI dependency on only two spikes makes it susceptible to noise, whereas spike count measures signals over a certain time window, and therefore mitigates variation in response caused by exact spike timing.  

\subsection{\textbf{Real-world data}} 
To study the performance of the TDE-2 and TDE-3 with real-world data, we used a dataset from \cite{rueckauer2016}. Specifically, we focused on sequence with boxes moving to the right (Figure \ref{fig:Figure11}) and disk rotating counter-clockwise (Figure \ref{fig:Figure12}). The former is an example of textured stimuli with multiple edges moving in the same direction and thus directly tests the robustness of TDE direction-selectivity and gains obtained by augmenting the TDE-2 with an inhibitory input. The latter contains a high dynamic range of velocities both in magnitude (80-fold) and direction (360\textdegree) and is used to probe velocity coding by TDEs in real-world scenes.

The real-world data was used as a test dataset. TDEs were pre-trained on synthetic data (see Sec. \ref{subsec:real_network}) using spike count as the inference method because it gives better performance in the presence of noise (Figure \ref{fig:FigureNew}). We pooled together ON and OFF events as it is often encountered in biological systems \cite{clark2016parallel,borst2020fly,sterling2015}.

\begin{figure}[h]
    \begin{minipage}{0.45\textwidth} 
        \centering
        \includegraphics[width=.9\linewidth]{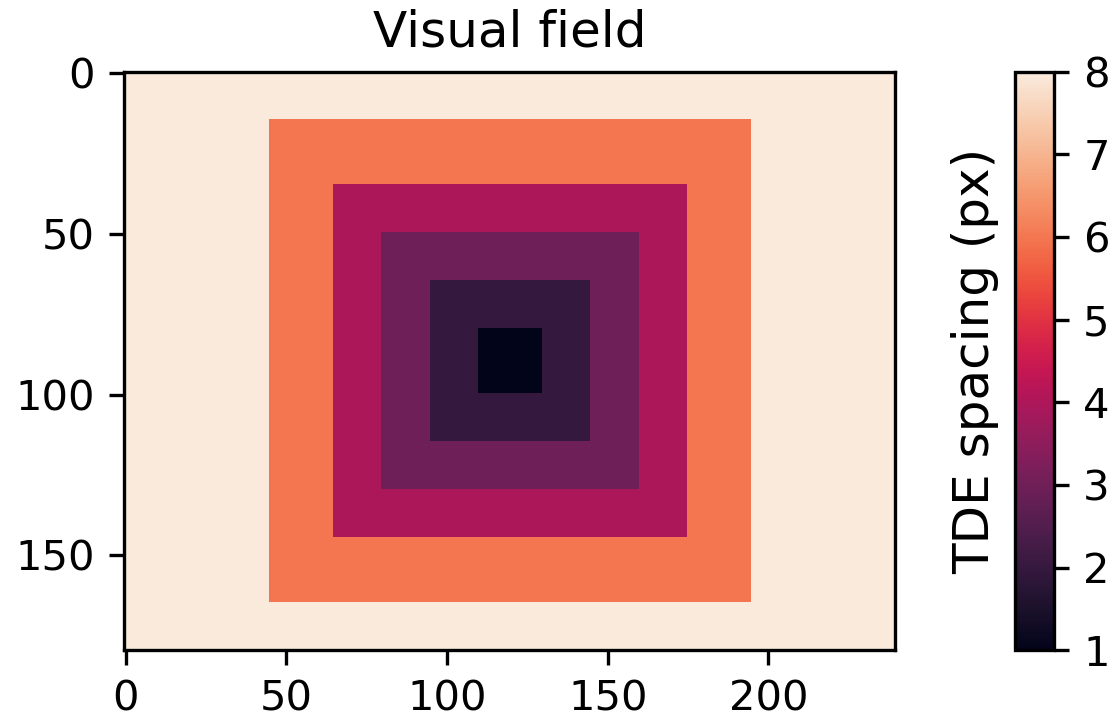} 
        \captionsetup{font={small, sf, it}} 
        \caption{Spacing between compartments of TDE as a function of its location within the visual field.}
        \label{fig:Figure4}
    \end{minipage}\hfill 
    \begin{minipage}{0.45\textwidth} 
        \centering
        \includegraphics[width=0.6\linewidth]{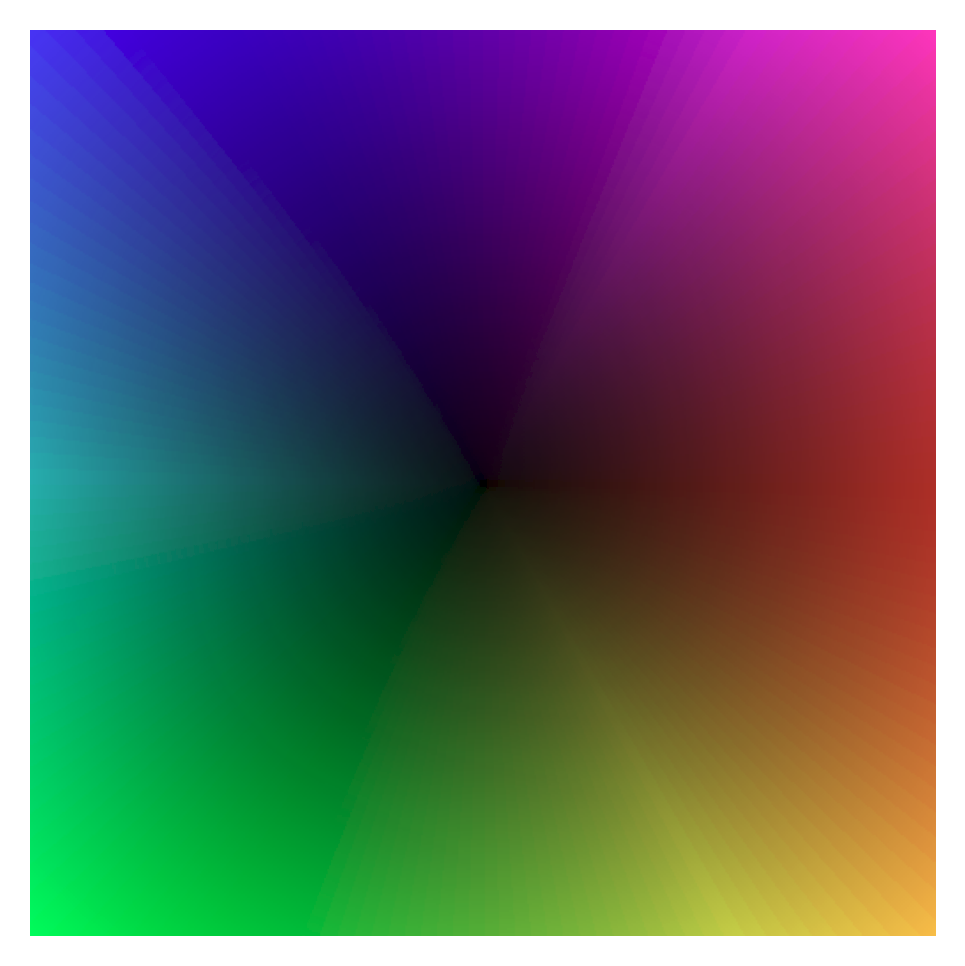} 
        \captionsetup{font={small, sf, it}} 
        \caption{Color map employed for the visualization of the optical flow. Direction is encoded with hue, magnitude with brightness.}
        \label{fig:Figure5}
    \end{minipage}
\end{figure}

\subsubsection{\textbf{Translating boxes}}

\begin{figure*}[h]
  \centering
  \includegraphics[width=1.0\linewidth]{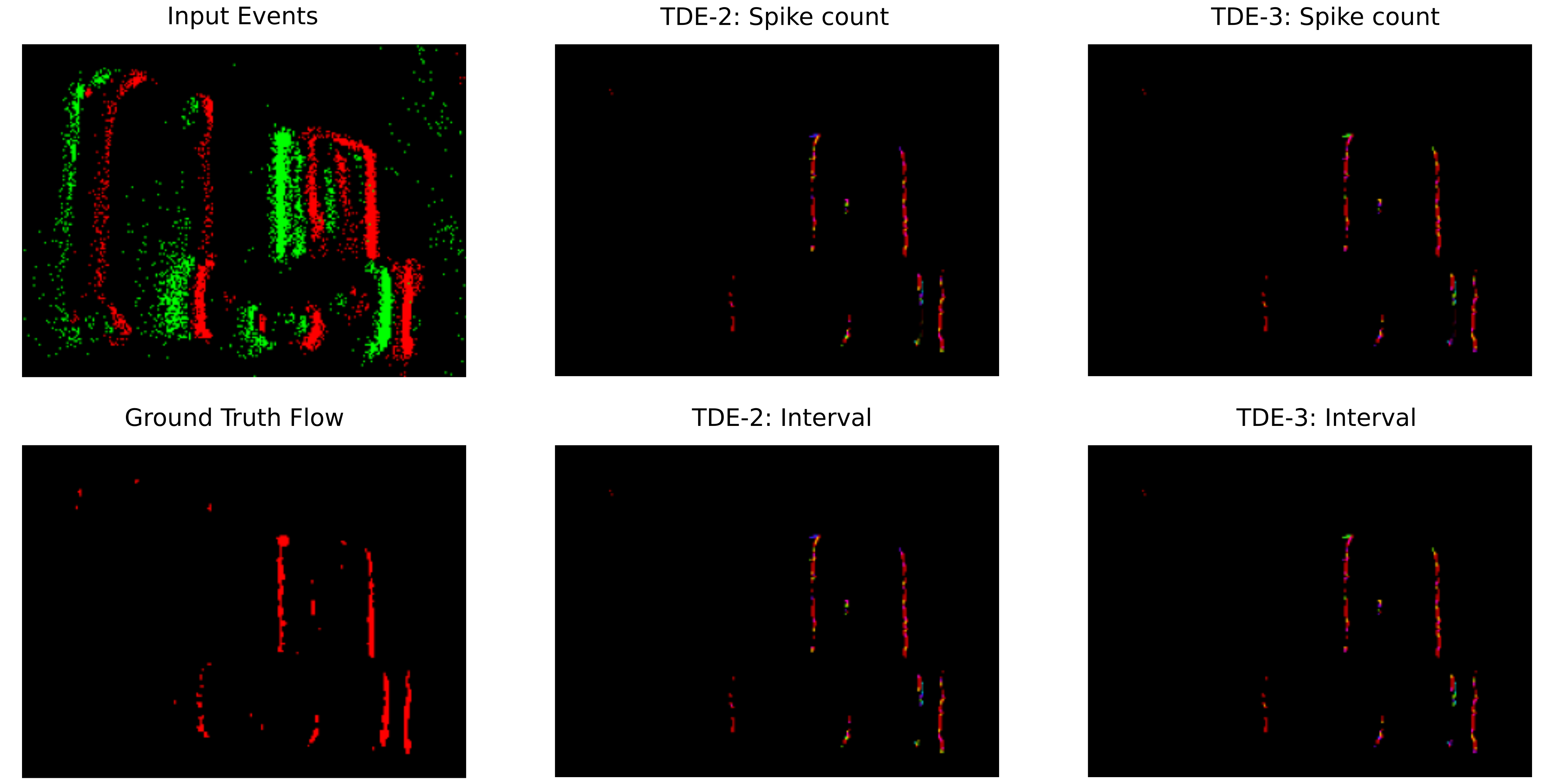}
  \caption{Optical flow coding by TDE: translating boxes. Raw events (top left panel) are color-coded as green-ON events, red - OFF events.  Optical flow (the rest of the panels) is color-coded according to Figure \ref{fig:Figure11}.}
  \label{fig:Figure11}
\end{figure*}

We start the analysis of TDE's responses to boxes moving to the right (Figure \ref{fig:Figure11}) by estimating the direction-selectivity of individual TDEs in a manner similar to DSI (Figure \ref{fig:Figure6}). 
Although strictly speaking, here we cannot calculate the DSI because the motion of the boxes was confined to one direction, we can approximate it by calculating the fraction of total spikes in the network that belonged to the detectors tuned in the L-R direction. Ideally, this number should be one, as only L-R detectors should spike. Table \ref{tab:boxes} shows that for the network of TDE-2 only 43\% of spikes are from the L-R detectors, while for the network of TDE-3 70\% of spikes are attributed to L-R detectors. Thus, just as in the case of synthetic data  (Figure \ref{fig:Figure6}), the TDE-3 has more robust direction-selectivity.

Next, we visualized optical flow coding by the TDEs using the colormap from Figure \ref{fig:Figure5} on Figure \ref{fig:Figure11}. 
Surprisingly, although the TDE-3 has a more robust direction-selectivity on the level of individual detectors, on the network level the TDE-2 and TDE-3 seem to have similar qualitative performance regardless of inference methods. 
This qualitative observation is also confirmed quantitatively: Table \ref{tab:boxes} indicates that in all of the testing conditions, the Average Angular Error (AAE)  ($\approx\ 20$\textdegree) and standard deviation of the angular error ($\approx\ 30$\textdegree) were almost identical. 

\begin{table}[H]
   \caption{TDE performance with real-world data: translating boxes. TDEs were trained with spike count inference (see Sec.\ref{subsec:real_network}} 
   \label{tab:boxes}
   \small
   \centering
   \begin{tabular}{lcccr}
   \toprule\toprule
   \textbf{Detector Type} & \textbf{Inference Method} & \textbf{mean AAE $\pm$ std (\textdegree) } & \textbf{Fraction of spikes in PD} & \textbf{Total N Spikes} \\ 
   \midrule
   \multirow{2}{*}{TDE-2} & spike count & 19 $\pm$ 31 & \multirow{2}{*}{0.43} & \multirow{2}{*}{229686} \\
                                  & ISI & 18 $\pm$ 31 & \\
   \midrule
   \multirow{2}{*}{TDE-3} & spike count & 18 $\pm$ 32 & \multirow{2}{*}{0.7} & \multirow{2}{*}{126692} \\
                         & ISI & 17$\pm$ 31 & \\
   \bottomrule
   \end{tabular}
\end{table}

The apparent contradiction between responses on the level of individual TDEs and the network level stems from the fact that to estimate and visualize optical flow, one needs to subtract signals of opposingly-tuned detectors from each other (Section \ref{subsec:real_network}). 
Now, the motion of an edge to the right leads to residual gain activity in T-B, B-T, and R-L tuned detectors which can be converted to TDE output upon the appearance of a second edge moving to the right (Figures \ref{fig:Figure1} and \ref{fig:Figure6}, Section \ref{subsec:TDE}). However, for B-T and T-B tuned detectors, this activation should be equal (except for noise), whereas activation of L-R tuned detector would be always higher than activation of R-L tuned detectors. Consequently, the subtraction of velocities estimated by the opposingly tuned detectors compensates for poor direction-selectivity of individual TDE-2s by removing signals caused by residual gain activity from the data stream.

Though subtraction of velocity estimates made by opposingly tuned detectors mitigates the loss of direction-selectivity of TDE-2 in a textured real-world environment, it is still better to employ TDE-3, because the inhibitory input eliminates residual activity immediately at the level of motion detection. As a result, the network becomes more computationally efficient as it spikes almost two times less (Table \ref{tab:boxes}), and therefore will be more energy-efficient when implemented in neuromorphic hardware \cite{caccavella_etal23}. To make a statistical argument for the higher energy efficiency of TDE-3 we also compared detector spike count with an additional 10 real-world scenes (rotating disk + 9 scenes from \cite{burner2022evimo2}) and found that in all of the visual scenes, TDE-2 spiked more than TDE-2 (on average $2.6 \pm 0.75$.  times,p=1e-4; see Supplementary Information Figure S2 for the details). 

To conclude, networks of TDE-2s and TDE-3s can both reliably infer motion direction in natural scenes with spike count and ISI-based inference. Yet, more robust direction selectivity of individual TDE-3 leads to better energy efficiency of the entire network.

\subsubsection{\textbf{Rotating disk}}

\begin{figure*}[h]
  \centering
  \includegraphics[width=1.0\linewidth]{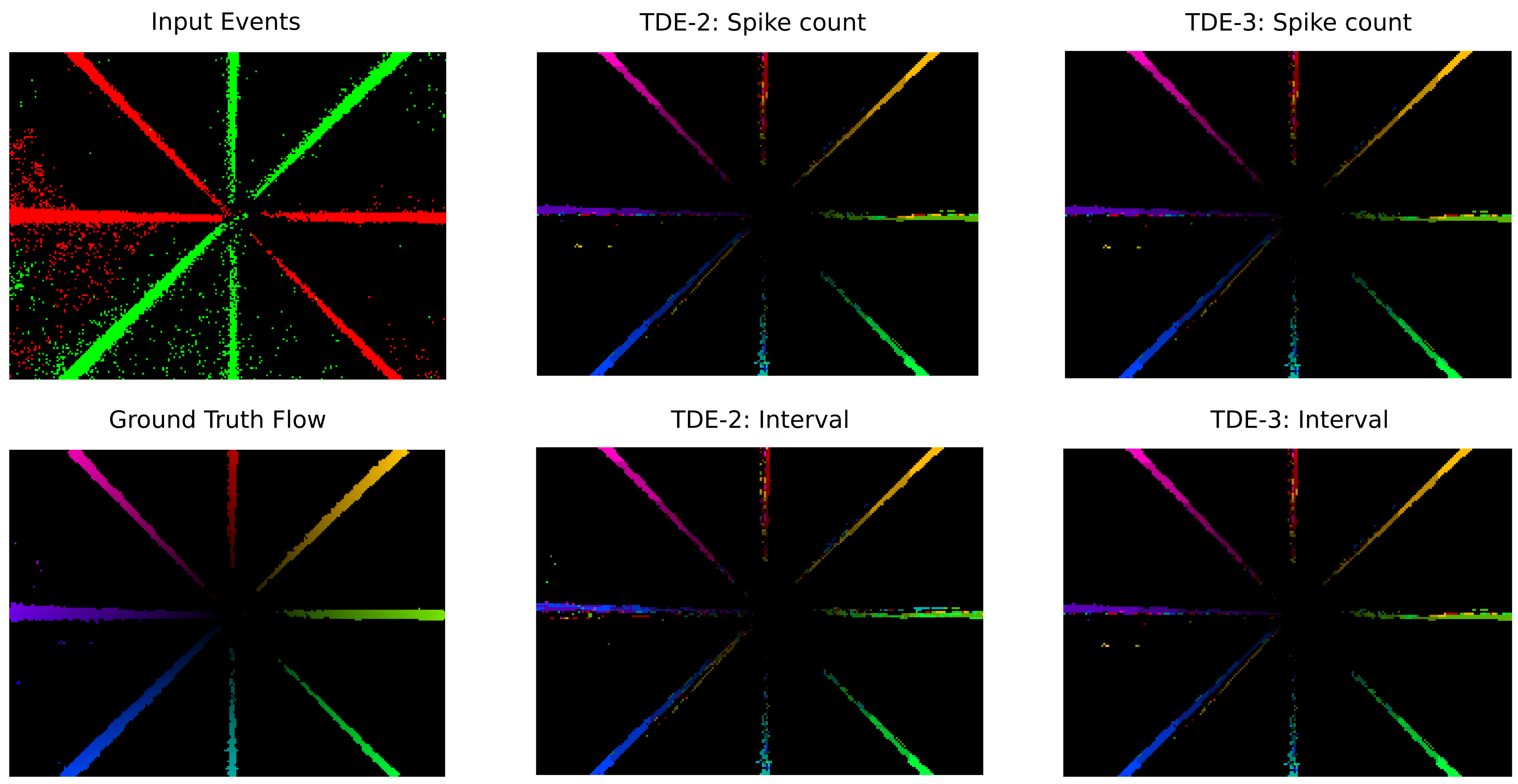}
  \caption{Optical flow coding by TDE: rotating disk.}
  \label{fig:Figure12}
\end{figure*}

TDEs ability to encode motion in a wide dynamic range of direction and magnitude was tested with the aid of a rotating disk sequence \cite{rueckauer2016} (Figure \ref{fig:Figure12} top panel) that challenged networks of TDEs with 360\textdegree range of motion direction and 80-fold range of velocity magnitude. 
To deal with such stimuli conditions, the network of TDEs tuned to 4 cardinal directions sampled space anisotropically, with the lower distance between TDE inputs in the center and higher in the periphery (Figure \ref{fig:Figure4}). This bioinspired approach \cite{dangelo2020event} allowed us to reconcile dynamic range, resolution, and latency requirements (see Section \ref{subsec:TDE}, \ref{subsec:training}). We assessed optical flow coding by networks of TDE-2 and TDE-3 with spike count and ISI-based inference. 

\begin{table}[H]
   \caption{TDE performance with real-world data: rotating disk.} 
   \label{tab:disk}
   \small
   \centering
   \begin{tabular}{lccccr}  
   \toprule\toprule
   \textbf{Detector Type} & \textbf{Inference Method} & \textbf{mean AAE $\pm$ std (\textdegree)} & \textbf{rAEE $\pm$ std} & \textbf{Correlation, r} & \textbf{Total N Spikes} \\ 
   \midrule
   \multirow{2}{*}{TDE-2} & spike count & 21 $\pm$ 24 & 0.43 $\pm$0.23 & 0.87 & \multirow{2}{*}{1638878}  \\
                                  & ISI & 21 $\pm$ 24 & 0.47 $\pm$ 0.18 & 0.88&\\
   \midrule
   \multirow{2}{*}{TDE-3} & spike count & 22 $\pm$ 24 & 0.45 $\pm$ 0.26 & 0.87 & \multirow{2}{*}{1532557}  \\
                                  & ISI & 21 $\pm$ 24 & 0.47 $\pm$ 0.2 & 0.88&\\
   \bottomrule
   \end{tabular}
\end{table}

Figure \ref{fig:Figure12} employs a color-coding scheme from Figure \ref{fig:Figure5} to visualize optical flow coding by various networks of TDEs to compare with ground truth (Figure \ref{fig:Figure12} left bottom). The Figure indicates that in all 4 testing conditions, there is a close correspondence between TDE inference and ground truth. This observation is also supported quantitatively. In all four cases, TDE showed similar (and fairly low, see \cite{rueckauer2016}) AAE (mean and std of $\approx\ 20$\textdegree, Table \ref{tab:disk}) and relative average endpoint error (rAEE, mean of $\approx\ 0.4$, std of $\approx\ 0.2$).

Although a 40\% error in velocity estimation might look substantial in terms of coding precision, one should keep in mind three things. 
First of all, we encode over a wide dynamic range of velocities. 
Secondly, each detector encodes velocities with few spikes/short ISI, so, naturally, such coarse binning would impair precision. 
However, most importantly, there is almost 90\% correlation between TDE velocity estimate and ground truth (Table \ref{tab:disk}). Indeed, optical flow depends on the ratio between velocity and distance, so it does not provide information about absolute velocity anyway \cite{SERRES2017703}. Meanwhile, a high correlation with velocity indicates that regardless of the exact scale, networks of TDE can reliably determine the relative distribution of the velocities in the visual scene.

To summarize, both TDE-2 and TDE-3 can extract a wide dynamic range of velocity direction in magnitude from real-world event-based data with both spike count and ISI-based velocity inference.


\section{Discussion}
In this work, we made a thorough investigation of the TDE as a computational prior for optical flow estimation in SNNs on three different levels: detector architecture, velocity inference, and real-world performance. Below we summarize the key outcomes of the study and discuss possible future improvements on each of these three levels. 

\subsection{Detector architecture}
On the level of detector architecture, we showed that classical TDE-2 loses direction-selectivity in a textured environment due to the residual activity in the gain compartment (Figures \ref{fig:Figure1}, \ref{fig:Figure6}, \ref{fig:Figure11}). We proposed a bio-inspired solution for this problem: to augment the TDE-2 with an inhibitory input that resets the gain and removes residual activity upon activation (Figures \ref{fig:Figure1}, \ref{fig:Figure6}, \ref{fig:Figure11}). This procedure preserves the direction-selectivity of individual TDE-3s (Figure \ref{fig:Figure6} and Table \ref{tab:boxes}) and increases the computational efficiency of the network of TDEs by decreasing the total number of spikes 2-4 times (Supplementary Figure S2), which decreases energy consumption proportionally when implemented on neuromorphic hardware \cite{caccavella_etal23}.

Although we obtained a principal result about the importance of inhibitory input for the performance of the TDE, there are two issues one needs to address to force TDE-3 to optimal performance: jitter in event timing \cite{Czech2016,Ding_2023} and the way inhibition is implemented. 

The jitter creates a problem for direction-selectivity because the reset of the gain during motion in the orthogonal direction is due to the simultaneous activation of facilitator and inhibitor compartments (See Sections \ref{subsec:TDE}, \ref{subsec:training}), which is hardly achievable with real-world data due to the jitter. A solution to this problem is to low-pass filter inputs to the facilitator and inhibitor compartments. Here it was done crudely by binning events into timesteps (which is still compatible with digital neuromorphic processors such as Loihi \cite{Davies_etal18}), but the TDE could benefit from the incorporation of an adaptive low-pass filter with trainable parameters. It is crucial for the filter to be adaptive because there is a trade-off between velocity resolution that requires short time constants and combating the noise that requires long time constants. Adaptability is crucial since the optimal time constant depends on the particular characteristics of the visual scene such as velocity distribution, illumination, contrast, spatial frequency, etc.

In the present study, inhibition was implemented as an instantaneous reset of the gain, modeled as a multiplication by zero. In contrast, in the insect elementary motion detector that inspired TDE-3, inhibition has its own dynamics and can influence neuronal resistance, current, membrane potential, and time constant \cite{borst2020fly}. These richer dynamics have mainly been studied in detailed biophysical models of insect vision \cite{borst2020fly}.

Notably, nearly two decades before 3-point elementary motion detectors were discovered in insects, a similar triplet architecture was proposed for neuromorphic hardware by Kramer \cite{Kramer}. His FTI (facilitator-trigger-inhibitor) detector, as well as related approaches \cite{Ros99,Higgins99,Deutschmann99}, used inhibition to suppress spiking by acting directly at the output. While TDE-3 shares the same functional goal — suppressing responses to motion in the non-preferred direction — it implements inhibition by resetting the gain, thereby intervening earlier in the computation. Other applications of triplet correlations for optical flow estimation from event-based data \cite{Shiba23spl} used simple averaging of velocity estimates between triplet pairs and did not rely on spiking neural networks or trainable systems. Grayscale-based detectors \cite{Bahroun2019,Mano2021} modeled inhibition as a linear subtraction between the activations of the inhibitor and facilitator compartments. These diverse implementations highlight the flexibility of triplet-based motion detection architectures.

In the present study, inhibition was implemented as an instantaneous multiplication of the gain by zero. However, in the insect elementary motion detector, which serves as inspiration for TDE-3, inhibition has its own dynamics and influences neuronal resistance, current, membrane potential, and time constant \cite{borst2020fly}. Nevertheless, such complex interactions have primarily been studied in the context of detailed biophysical models of insect motion detection \cite{borst2020fly}. Notably, nearly two decades before 3-point elementary motion detectors were discovered in insects, similar architectures had already been proposed for neuromorphic hardware by Kramer \cite{Kramer} from a purely engineering perspective. In his FTI (facilitator-trigger-inhibitor) detector, as well as in related approaches \cite{Ros99,Higgins99,Deutschmann99}, the inhibitor acted directly on the neuronal output, preventing firing upon activation. Other applications of triplet correlations for optical flow estimation from event-based data \cite{Shiba23spl} employed a simple averaging of velocity estimates between each pair within the event triplet and did not use SNNs or any type of neural network. Studies using grayscale images \cite{Bahroun2019,Mano2021} implemented detectors where inhibition was modeled as a simple subtraction of the inhibitor's activation from that of the facilitator.

Future research could benefit from exploring more biologically accurate inhibition dynamics and systematically investigating different interaction modes between TDE-3 compartments.

\subsection{Velocity inference}

For the velocity inference, we developed a procedure for supervised training of the TDEs to linearly encode input velocities using spike count and ISI-based inference (Figures \ref{fig:Figure2},\ref{fig:Figure3}). For the ISI-based training, we developed a novel approach. While most other methods \cite{BOHTE200217, Hesham, Comsa} model the relationship between input magnitude with differentiable function than use BPTT, we measure ISI from the amplitude of low-pass filtered spike train to use BPTT with surrogate gradient (see Section \ref{subsec:isi} for the details). The theoretical advantage of our method is that it does not require estimation of the relationship between spike timing and input magnitude, allowing work with more conventional activation functions. 

We investigated the performance of both inference methods in conditions with wide (Figure \ref{fig:Figure7}) and narrow (Figure \ref{fig:Figure8}) dynamic ranges of input velocities and tested their robustness towards variation in stimulus spatial frequency (Figure \ref{fig:Figure9}) and background activity noise (Figure \ref{fig:Figure10}). 
We found that although in a noise-free environment, ISI-based inference performs slightly better than spike count in terms of the correlation coefficient between true and estimated velocity (Figures \ref{fig:Figure7}, \ref{fig:Figure8}, \ref{fig:Figure9}) when the noise is present spike count turns out to be more reliable, especially when both training and test inference are count based (Figures \ref{fig:Figure10}, \ref{fig:FigureNew}). 

Overall, we found that the TDE-based networks are robust in a wide range of noise conditions, showing FTA of >95\% and correlation with stimulus velocity of $\approx 90\%$ at all noise levels except the two highest (5 Hz/px and 10 Hz/ px, Figure \ref{fig:FigureNew}). However, we suspect that in those conditions it is possible to leverage efficient TDE performance. To this end, we suggest more aggressive filtering at the STCF stage, as was done for the real-world data (Figures \ref{fig:Figure11}, \ref{fig:Figure12}). 

Examiming training progression we observe that while the loss curves for training with spike count inference tend to be relatively smooth, the loss curves for training with ISI exhibit noticeable jumps, with changes in the loss value appearing more pulsatile than gradual. This behavior can be attributed to the initialization, which results in very high spiking rates (see Section \ref{subsec:training} for the details). Under these conditions, it is easier to regulate spike count than the ISI (inter-spike interval) between the first two spikes. For spike count, improving the loss function requires only a slight adjustment in the spike count for one of the stimulus velocities. In contrast, ISI regulation is more challenging because the ISI tends to decrease over time (being shortest immediately after trigger activation). Consequently, training with ISI requires more epochs to produce meaningful differences in velocity estimation that also manifest themselves in more abrupt fashion.

To decode stimulus velocity from TDE activity in the presence of multiple edges one needs to perform motion segmentation to relate counted spikes/measured ISIs to a particular moving edge. How to know when to count spikes and to measure ISI? Here we proposed to count spikes starting from each activation of TDE's trigger as it indicates passage of a new edge. This worked for a shallow network where we could read out trigger activations, but suppose there would be more layers after the TDE. How could downstream circuitry perform motion segmentation? One obvious solution would be to establish so-called skip connections between input to TDE's trigger and downstream circuitry. Another solution could exploit the fact that immediately after activation of the trigger the TDE's signal is maximum i.e. if one would count spikes within a certain sliding time window, the local (in time) maximal count would be close to the moment of edge appearance at TDE's trigger. Similarly, the shortest distance between two spikes would be for the first two spikes after trigger activation. Hence, downstream circuitry potentially can perform motion segmentation by performing temporal max pooling.  

For the velocity coding, the fundamental limitation of the TDEs is the latency vs. dynamic range dilemma (see Section \ref{subsec:TDE}). Since spike amplitude is usually fixed, high dynamic range requires a lot of spikes (or timepoints between spikes), which necessarily leads to a long latency in velocity inference. 
The best way to solve this problem is to implement TDE on the neuromorphic chip Loihi2 (\cite{orchard2021efficient,Chiavazza_etal23} that allows "graded" spikes of different amplitude. As a result, one does not need to count spikes or intervals between spikes, because velocity magnitude is immediately available from the first spike. However, apart from Loihi2, no neuromorphic hardware supports spikes of different amplitude.  Therefore, one could also solve the latency vs. dynamic range dilemma by employing the so-called "eccentric downsampling" using several detectors tuned to different velocity bandwidths such that each detector emits only a few spikes (Figure \ref{fig:Figure4} \cite{dangelo2020event}).

\subsection{Performance with real-world data}

To the best of our knowledge, our experiments on Figures \ref{fig:Figure11},\ref{fig:Figure12} are the first to \textbf{evaluate} velocity coding with TDEs in real-world data.\footnote{A preprint by Greatorex et al. \cite{Chicca25} presenting an ego-motion estimation with a network of TDE-2 appeared on arXiv shortly after the initial release of our manuscript}. Comparison between TDE-2 and TDE-3 performance with the real-world data shows that despite higher energy efficiency of TDE-3 network (see Table \ref{tab:boxes}, Supplementary Figure S2), both detectors perform almost equally in terms of determination of motion direction (AAE, Table \ref{tab:boxes}, \ref{tab:disk}) and motion velocity (AEE, Table \ref{tab:boxes}) across 360\textdegree of change in direction and the 80-fold change in velocity magnitude (Figure \ref{fig:Figure12}). TDEs performed with precision similar to model-based elementary motion detectors \cite{rueckauer2016}, but have the advantage of being naturally compatible with SNNs and thus can be implemented in low-power neuromorphic hardware to exploit the spatio-temporal sparsity of event-based streams.

Subtraction of velocity estimates obtained from opposingly tuned detectors improves TDE velocity coding on the network level and seems like a very intuitive post-processing step. Yet, at this level, we still deal with very shallow networks. Future work should aim at the incorporation of TDE-3 into deeper spiking networks with learnable interactions between individual TDE-3 and the interplay between spike count and ISI-based inferences. Deeper networks would also allow one to mitigate drawbacks (such as latency vs. dynamic range dilemma) and highlight the benefits of the TDE-3.

Our experiments with real-world data suggest the training of the TDEs is generalizable. Indeed, TDE-3 was trained in simulation and yet worked fine in the real world. Moreover, we trained a single detector, expressing velocities in abstract quantities of velocities, expressing them as a fraction of the distance between facilitator and trigger covered in one algorithmic timestep (see Section \ref{subsec:real_network}) and subsequently scaled the dynamic range of encoded velocities by simply changing the distance between facilitator and trigger. Thus, due to TDEs simplicity, one can easily adapt it to a variety of conditions with an appropriate choice of spacing between TDE outputs and the duration of the timestep. 

We hypothesize that TDE-3 can enhance the generalizability of SNNs by extracting simple, abstract features—such as moving edges—that are ubiquitous in visual scenes. In support of this, research on drone acrobatics has shown that autopilots trained in abstract representations of visual scenes bridge the simulation-to-reality gap more effectively than those trained on raw images \cite{kaufmann2020deepdroneacrobatics}. We believe that for low-level visual processing, TDE-3 offers greater robustness compared to Convolutional Neural Networks (CNNs). Unlike CNNs, which rely on architectures shaped by training data and are therefore prone to overfitting, TDE-3 employs a fixed, analytic architecture, reducing the risk of overfitting and enhancing stability.

To conclude, we presented in this paper the TDE-3 as a potent computational prior for motion detection from event-based data in SNNs. It can be trained and allows fast, efficient, precise, and robust coding of optical flow. In a broader sense, our study highlights the importance of understanding the relevant biological features (e.g. relation between structure and function) in the design of neuromorphic solutions for real-time and low-power motion detection.
\section{Methods}
\label{sec:Methods}

\subsection{\textbf{Datasets}} \label{subsec:data}

\subsubsection{\textbf{Synthetic data}}
The synthetic dataset consists of vertically oriented bars of three light intensity values: white, grey, and black. The greyscale values were converted into events using a simplistic custom-made simulator of an event-based camera, which retained its key property: encoding changes in greyscale value with spiking signals that report increases (on events) or decreases (off events) in light intensity. Specifically, we modeled voltage in the event-based detector as being proportional to the logarithm of light intensity and reported contrast each time the temporal derivative of this signal \((V_t - V_{t-1})\) was higher/lower than the threshold value of \(\pm 0.15\).

To compare the robustness in direction-selectivity of TDE-3 and TDE-2, we simulated their responses to moving textured stimuli consisting of bars oriented orthogonally towards the direction of motion. The bars had three intensity levels: white, grey, and black. The size of the texture along the motion axis was 80 pixels and 3 pixels along the orthogonal direction (Figure \ref{fig:Figure6}A). We employed 5 velocities: 0.1 px/timestep, 0.2 px/timestep, 0.33 px/timestep, 0.5 px/timestep and 1 px/timestep. The motion direction (left-right, right-left, top-bottom, bottom-top) and velocity were randomly chosen for each stimulus example (2000 per testing round). To vary the "amount" of texture in stimuli we randomly varied the fraction of the grey bars from 0\% to 80\%. 

To train the TDE-3 to linearly map velocities of moving edges into spike count-based or ISI-based output, we employed isolated edges on a homogeneous background. To avoid over-fitting and the necessity to separate stimuli into training and test datasets, for each training epoch we de-novo randomly generated 100 examples of moving edges with various velocities. We employed two sets of velocities. One was relatively wide (5 velocities spanning a 10-fold range, see above). The other was relatively narrow with 15 velocities within a range from 0.025 to 0.04 px/timestep. This stimulus set was employed to show that the TDE can encode velocities with high precision. 

To study the effect of variation in scene spatial frequency on the coding precision of ISI- and spike count-based inference, we trained the TDEs to linearly encode velocities of two edges moving at the same velocities at different distances from each other. We used 5 velocities from 0.1 to 1 px/timestep and randomly varied the distance between edges within a range of 3,4,5,7 and 10 pixels.

To study the effect of noise on training and inference with spike count and ISI-based methods, we simulated one of the most prominent types of noise in event-based vision -  background activity noise, as a Poisson process following the approach by \cite{guo2023}. The noise was injected after the conversion of greyscale values to events as some stochastic events. We used six mean rates of noise: 0.2 Hz/px, 0.5 Hz/px, 1 Hz/px, 2 Hz/px, 5 Hz/px, and 10 Hz/px. This effectively covered the dynamic range in event-based vision from high to low SNR conditions \cite{guo2023}. For this set of experiments, we employed the same set of 5 velocities as above and varied the distance between edges within a range of 3 to 10 pixels as above. The timestep duration was set to 10ms. 

\subsubsection{\textbf{Real-world data}}
We used real-world data from the dataset provided by \cite{rueckauer2016}. The data was collected with an event-based camera DAVIS 240C with a resolution of 240*180 pixels and contained various sequences of the full-field motion with a duration between 1 second and 4 seconds.
We were specifically interested in two sequences: one with the apparent motion of the various boxes to the right (Figure \ref{fig:Figure11}) and the one with the disk rotating clockwise (Figure \ref{fig:Figure12}). The former sequence contained multiple moving edges closely following each other with very similar velocities and thus directly targeted the key weakness of the TDE-2 of loss in direction-selectivity in a textured environment (Figures \ref{fig:Figure1} and \ref{fig:Figure6}. The second sequence contained a large dynamic range of velocities (proportional to the distance from the center of rotation) and was thus ideally suited to test the optic flow coding properties of the TDEs.

The dataset contained ground truth optical flow that could be inferred from the IMU measurements. Indeed, motion in the sequences was confined to the motion of the camera: for the boxes to yaw and for the disk to roll rotations (z-axis). Hence, in this case, ground truth optical flow can be obtained as a displacement vector of individual events from the IMU measurements with equation \ref{eq:equation1} following \cite{rueckauer2016}:

\begin{equation} \label{eq:equation1}
\mathbf{e'} = \mathbf{R}(\mathbf{e} - \mathbf{e_0} + \mathbf{T}) + \mathbf{e_0}
\end{equation}
where $\mathbf{e}$ is the initial location of an event $(e_x, e_y)$, $\mathbf{e'}$ - end location of the event, $\mathbf{e_0}$ is the location of IMU center, $\mathbf{T}$ describes motion due to yaw and pitch rotations, and $\mathbf{R}$ describes motion due to roll rotation (z-axis):
\begin{equation} \label{eq:equation2}
\mathbf{T} = k \begin{bmatrix}
    y \\
    x
\end{bmatrix} ; \quad \mathbf{R} = \begin{bmatrix}
    \cos z & -\sin z \\
    \sin z & \cos z
\end{bmatrix}
\end{equation}

where x,y, and z are the rotation rates of IMU around corresponding axes (pitch, yaw, and roll rotations) and k is a scaling factor that converts degrees to pixels for the yaw and pitch rotations equal to 4.25 px/deg \cite{rueckauer2016}.

\subsection{\textbf{Inference and training}}\label{subsec:training}

\subsubsection{\textbf{Training and inference with the synthetic data}}
The SNNs were simulated using the PyTorch framework, inspired by previous implementations \cite{Neftci_etal19,hagenaars2021selfsupervised}. Depending on the application, networks included several types of elements such as TDEs, STCFs, and low-pass filters of the spike trains (i.e. spike traces). Each element was defined by its synaptic weights and time constants. The TDE trainable parameters included gain, current, and voltage time constants and synaptic weight of the gain compartment. To enhance learning of the neuronal time constants, we used the approach developed by \cite{fang2021incorporating} that reformulated neuronal decays as the result of multiplying gain, current, and voltage with the Sigmoid of a trainable parameter.

\begin{figure}[h]
    \centering
    \begin{tikzpicture}[node distance=4.2cm, auto]
        \begin{scope}[scale=1.0, transform shape]
            \node (input) [startstop] {Input};
            \node (sptcf) [io, right of=input, align=center] {Spatio-Temporal\\
                                                              Correlation Filter  };
            \node (tde) [io, right of=sptcf] {TDEs};
            \node (vel) [process, right of=tde] {Velocity = $\Sigma$ spikes};
            \draw [arrow] (input) -- (sptcf);
            \draw [arrow] (sptcf) -- (tde);
            \draw [arrow] (tde) -- (vel);
        \end{scope}
    \end{tikzpicture}
    \captionsetup{font={small, sf, it}, width=1.0\linewidth} 
    \caption{Pipeline for supervisory training of TDE with BPTT to linearly map velocities with spike-count based inference}
    \label{fig:Figure2}
\end{figure}

To train SNNs with BPTT, we tackle the non-differentiability problem of the spiking activation function by using surrogate gradients \cite{Neftci_etal19, zenke2021remarkable}. The method enables backpropagation in SNNs by replacing the spiking activation function with a differentiable function with a similar shape and steep slope during the backward pass \cite{Neftci_etal19,friedemann_zenke_2019_3724018,zenke2021remarkable}. We used the surrogate gradient proposed in \cite{friedemann_zenke_2019_3724018}.

To mitigate the effect of the background activity noise in experiments with real-world data and injected noise, we employed STCF filtering described by  \cite{guo2023}. In this simple yet efficient algorithm, an event is allowed to pass the filtering stage only if there is at least 1 event in n nearest neighboring pixels (trainable parameter) within a specified time window. Essentially, the filter extracts correlations (hence the name) between a pixel and its neighborhood while rejecting uncorrelated noisy events. As a result, STCF improves the signal-to-noise ratio while preserving signal sparsity.

To promote linear mapping of input velocity into velocity estimated from TDE-3's output activity, we used the mean absolute error (L1 loss) between the normalized TDE-3 estimates and the ground truth edge velocities. Normalization ensures linear mapping because it removes scale such that the loss depends only on the non-linearities of the mapping. Normalization was done with respect to the maximum estimate/velocity in the batch. Additionally, to ensure the sparsity of the learned solution, we added a regularization loss which is proportional to the mean squared spike count. At the same time, to avoid gradient vanishing due to the low neuronal activity \cite{Rossbroich_2022}, the network was initialized to have a high firing activity by making synaptic weights large and time constants long. 

Depending on the coding scheme (spike count or ISI), we used two slightly different training pipelines to extract motion direction and velocity (Figures \ref{fig:Figure2},\ref{fig:Figure3}). The two first steps are common to both pipelines. The input layer simply converts events to spikes. Then, these spikes are fed into the layer of TDE-3s, which effectively perform 1D convolution with kernel size=3, stride=1, and weights being facilitator, trigger, and inhibitor. This layer consists of two channels, one for each motion direction. At the next processing stage, spike count-based and ISI-based pipelines deviate. For the spike count-based velocity estimator we can straightaway find the total number of spikes emitted by each of the detectors, convert it to estimated velocity, calculate the loss, and perform a backward pass. For the ISI-based velocity estimation, the training of the network is not so straightforward and is described in section \ref{subsec:isi}.

For the velocity set with a wide (10-fold) dynamic range of velocities, the detector is trained to linearly map this dynamic range into limited spiking output. Hence it makes sense to estimate velocity as a spike count scaled by minimal discernible velocity (0.1 px/timestep), such that minimal velocity is encoded with one spike. For the narrow dynamic range (1.5-fold), the detector is trained to encode a limited dynamic range with very high precision. Hence, it makes sense to use a bias factor: to infer velocity, we multiply the spike count by the velocity resolution (0.001 px/timestep), and then, if the spike count is larger than 1, we add to it a bias factor of 0.024 px/timestep. This bias factor is necessary because it allows encoding a velocity of 0.025px/timestep with only 1 spike. Without this bias factor to have a velocity resolution of 0.001 px/timestep, would have required to use of 25 spikes for the velocity of 0.025 px/timestep.

To assess TDE-3 performance in the presence of multiple moving edges, one needs to perform motion segmentation: to relate counted spikes to particular edges. To do so we (1) limit the temporal window within which spikes are counted to 10 timesteps and (2) count spikes starting from the increase in TDE current. An increase in current is always a consequence of activation of the trigger. This means that the motion of a new edge causes it and thus solves the problem of motion segmentation. When we trained the detector to encode velocity amidst variation in stimulus spatial frequency, we calculated the error for the velocity estimate of each of the edges. 

A series of fast-moving edges can result in the integration of a very large signal in the current compartment of TDE-3, which would lead to saturation of TDE-3 spiking output. To prevent this situation, we clamped the maximal amplitude of the current compartment to the synaptic weight of the facilitator input. 

In the experiments with noise (Figures \ref{fig:Figure10}, \ref{fig:FigureNew}), not every increase in current is caused by a moving edge, and one of the goals of the training is to suppress the TDE activations caused by noise. To account for this factor in the calculation of the loss, we calculated the loss between all of the velocity estimates performed by the TDE and stimulus ground truth at corresponding timepoints (effectively, this ground truth was zero everywhere except at the moment of edge appearance see Section \ref{subsec:data}). Also, to evaluate how the trained TDE can reject noise and preserve the true motion signal, we employed an additional metric called fraction of true activity (FTA) as the sum of estimated velocities caused by the stimulus (true motion signal) to the total sum of estimated velocities (caused by signal and noise).

\subsubsection{\textbf{Training of the spiking neural networks to code with interspike interval}} \label{subsec:isi}

Apart from spike count, another intuitive method to encode signals in SNNs is relative spike timing \cite{Gollisch,Rieke99}. The main benefit of such signal coding is that it allows for much shorter latency \cite{Rieke99,Gollisch} while also enhancing coding capacity on a network level \cite{Paugam-Moisy2012}. Training of the SNNs to have spike timing has a long history, with the first classical paper being published as early as 2002 \cite{BOHTE200217}.
\begin{figure}
    \centering
    \begin{tikzpicture}[node distance=3 cm]
        \begin{scope}[scale=0.95, transform shape]
            \node (input) [startstop] {Input};
            \node (sptcf) [io, right of=input, align=center] {Spatio-Temporal \\
                                                              Correlation Filter};
            \node (tde) [io, right of=sptcf] {TDEs};
            \node (trace) [io, right of=tde] {Spike trace};
            \node (isi)   [io, right of=trace] {ISI};
            \node (vel) [process, right of=isi] {velocity $ \sim 1/\text{ISI}$};
            \draw [arrow] (input) -- (sptcf);
            \draw [arrow] (sptcf) -- (tde);
            \draw [arrow] (tde) -- (trace);
            \draw[arrow]  (trace)--(isi);
            \draw[arrow]   (isi)--(vel);
        \end{scope}
    \end{tikzpicture}
    \caption{Pipeline for supervisory training of TDE with BPTT to linearly map velocities with ISI-based inference.}
    \label{fig:Figure3}
\end{figure}

Here we developed a novel method to train SNNs to encode a signal with ISI using supervised learning with BPTT (Figure \ref{fig:Figure3}). While most other methods \cite{BOHTE200217, Hesham, Comsa} model the relationship between input magnitude with differentiable function than use BPTT, we measure ISI from the amplitude of low-pass filtered spike train to use BPTT with surrogate gradient. The theoretical advantage of our method is that it does not require estimation of the relationship between spike timing and input magnitude, allowing work with more conventional activation functions. 

Specifically, we were interested in the ISI between the first two spikes, as it should provide the shortest latency \cite{Milde,gutierrez-galan2022event}. The key is to estimate the time between two spikes in a differentiable manner. To do so we added to the pipeline on Figure \ref{fig:Figure2} low-pass filtering of the TDE output spikes (Figure \ref{fig:Figure3}).The filtering yields an exponentially decaying spike trace $x(t)$ (\ref{eq:equation3}):
\begin{equation}
\label{eq:equation3}
x(t) = x_0 \cdot e^{-\frac{t}{\tau}},
\end{equation}
where $x_0$ is spike trace amplitude upon arrival of a spike, $t$ - time lapsed since the arrival of the spike, and $\tau$ - filter time constant.

Solving equation \ref{eq:equation3} with respect to $t$ one can estimate this time $t$ with equation (\ref{eq:equation4}):
\begin{equation}
\label{eq:equation4}
t(x) = \tau \cdot \ln \left( \frac{x_0}{x} \right)
\end{equation}
Next, by taking filter output at the timestep directly preceding the second spike one can estimate ISI as $ISI=t(x)+1$. Then, by taking an inverse of it and scaling it by a proper factor ($\alpha$) one can convert it to the estimate of the input velocity \( v_{\text{est}} = \frac{\alpha}{\text{ISI}}\). Finally, to train a network with BPTT one can calculate the loss function using equation (\ref{eq:equation5}):
\begin{equation}
\label{eq:equation5}
L = \frac{1}{N} \sum_{i=1}^{N} |\hat{v}_{\text{est},i} - \hat{v}_{\text{true},i}|,
\end{equation}
where \( \hat{v}_{\text{est},i} \) and \( \hat{v}_{\text{true},i} \) represent the velocities for the \( i \)-th sample, and \( N \) is the total number of samples.

On the practical level, there are three factors to take care of to successfully implement a training pipeline based on the ISI:
\begin{enumerate}
    \item ISI interval is undefined when the TDE emits less than two spikes because the notion of interval requires 2 points. We assume that <2 TDE spikes mean an extremely slow motion of an object. Therefore we set ISI to a very large value ($10^3$ - $10^6$ timesteps) such that when inverted it will lead to an extremely low estimated velocity.
    \item Scaling of the inverted ISI to estimate velocities. This can be done similarly to scaling of spike count: for the stimuli set with a wide dynamic range we multiplied the maximal inverted ISI (=1) with maximal stimuli velocity; for the stimuli set with a narrow dynamic range, we also used bias and multiplied inverted spike count with the difference between maximal and minimal stimuli velocity.
    \item Handling of multiple moving edges. We opted to handle multiple moving edges similarly to spike count inference: the first two spikes were determined with respect to the current increase. Also,  In the presence of noise, the loss function and performance evaluation were handled similarly to the spike count. 
\end{enumerate}

\subsubsection{\textbf{Nuances of the application of TDEs to the real-world data}} \label{subsec:real_network}
Real-world data was used only as a test dataset.  We pooled together ON and OFF events as it is often encountered in biological systems \cite{clark2016parallel,borst2020fly,sterling2015}. The network was trained with synthetic moving bars to fit its dynamic range to the dynamic range of velocities in real-world data. Compared to simulation, estimation of the optical flow with TDEs in the real-world data has three additional challenges: (1) other (not only background activity) types of noise, (2) the necessity to coordinate the activity of multiple TDEs, and (3) the high dynamic range of velocities (2 decades for the rotating disk). 

Apart from background activity noise, another prominent source of noise in the event-based data is jitter in event timing \cite{Czech2016}. Event arrival is specified with microsecond precision, yet the time between actual change in intensity and event generation can vary in a range of milliseconds. Therefore, strictly speaking, each pixel of a moving edge generates an event at a different time \cite{Czech2016}. This is especially problematic for the 3-point TDE because the prevention of its activation upon motion in an orthogonal direction is underpinned by simultaneous activation of facilitator and inhibitor, which is hardly possible given the event jitter. To avoid this problem, one needs to increase the co-activation of the facilitator and inhibitor. The most straightforward way to do so is to downsample the event stream in time. For instance, feeding events to the network using time bins, where events within a bin are considered to occur simultaneously. There is a trade-off between mitigation of event jitter and dynamic range of velocity coding. Here (Figures \ref{fig:Figure11}, \ref{fig:Figure12}), given the relatively low stimulus velocity, we chose to slice events into bins of 50 ms (i.e. 20 frames per second). 

In contrast to synthetic data where we mostly employed either one or two opposingly tuned detectors, with the real-world data we sampled each pixel of the visual field with TDEs tuned along 4 cardinal directions: Left-to-Right (LR), Right-to-Left (RL), Top-to-Bottom (TB), and Bottom-to-Top (BT) akin to how it is done in vertebrae and insect visual systems \cite{clark2016parallel}. 
Although there are 4 cardinal directions, motion occurs along 2 axes: horizontal and vertical. Since a point in space cannot simultaneously move in 2 opposite directions, the next logical step in the processing pipeline is to subtract velocity estimates obtained from opposingly tuned detectors. This move is ubiquitously employed in biological systems \cite{borst2020fly,clark2016parallel,reichardt1961autocorrelation} and was shown to optimize velocity coding by biological elementary motion detectors \cite{KUHN2019963}. Here we found that such a subtraction compensates for the poor direction-selectivity of individual TDE-2.

The sequence with a rotating disk contained a very high dynamic range of velocities (2 decades: from 1 to 80 px/s). As was discussed earlier, using a single TDE to encode such a high dynamic range with high resolution is sub-optimal in terms of latency. Yet, it is possible to do so using a set of detectors tuned to different velocities. One elegant way to achieve so is to gradually increase spacing between TDE inputs (i.e. facilitator, trigger, inhibitor) as the detector position moves away from the center \cite{dangelo2020event} (Figure \ref{fig:Figure4}). Indeed, given internal parameters, the TDEs where the facilitator and trigger are spaced by n pixels can encode n times higher velocities than the detector where the facilitator and trigger are spaced by 1 pixel.

Although such an eccentric increase in TDE receptive fields nicely fits the rotation disk (where velocity increases as one moves out of the center), this solution is rather general and universally encountered in animal visual systems \cite{sterling2015}. The reason for this is that most of the optical flow is engendered by self-motion and most of the self-motion is directed forward. As animals tend to look towards the focus of expansion (a point in the visual field from which optical flow originates), the further an object is from the center of the visual field, the higher its retinal velocity. Given that robots also mostly move forward, eccentric increase in the receptive field is beneficial for robotics applications as well \cite{dangelo2020event}.

The specifics of our implementation are as follows. Using synthetic data, we first pre-trained TDE with spacing between compartments of 1 px to linearly encode 5 levels of velocity (from 0.1 px/timestep to 0.5 px/timestep). Now, it is important to recognize that the TDE encodes only the delay between activations of the facilitator and trigger compartments. Hence, the results of the training are indifferent towards the spatial configuration of the TDE, i.e. resultant velocity coding is not so much about px/timestep as about a fraction of the distance between facilitator and trigger covered within a timestep. Hence, by simply regulating the distance between the TDE inputs by a factor of n, one can scale the TDE range of velocity by said n without the need to re-train the TDE. Here, we used TDEs with distances between inputs of 1, 2, 3, 4, 6, and 8 pixels (Figure \ref{fig:Figure4}).  
Paired with a timestep duration of 50ms, it means that together the TDEs were able to cover velocities in the range of 2 to 80 px/s. For the spike count inference, the window over which spikes were counted was limited to 5 timesteps. Before feeding into the layer of TDEs signal was filtered by 3 by 3 STCF filter \cite{guo2023}.

\subsubsection{\textbf{Performance evaluation for the real-world data }}
For the qualitative comparison between the optical flow inferred with the TDEs and the ground truth, we visualized the optical flow using the colormap in Figure \ref{fig:Figure5}.
To make quantitative comparisons, we used the Angular Average Error (AAE), Average Endpoint Error (AEE), relative Average Endpoint Error (rAEE), and Pearson correlation coefficient \cite{Pearson1895}. AAE measured the difference between the direction of motion estimated with the TDEs and the ground truth motion direction, and it was calculated with equation \ref{eq:equation6}:

\begin{equation}
\label{eq:equation6}
AAE = \frac{1}{N} \sum_{i=1}^{N} \arccos\left(\frac{v_{ix}u_{ix} + v_{iy}u_{iy}}{|\mathbf{v}_i||\mathbf{u}_i|}\right)
\end{equation}

where \textbf{v} is the velocity vector estimated with the TDEs, \textbf{u} is the ground truth vector, and i denotes individual optical flow estimates (i.e. per pixel per timestep). Zero velocity measurements were not counted in Equation \ref{eq:equation5}. 
AEE was calculated as the difference between optical flow vectors inferred with TDE and ground truth using \ref{eq:equation7}:

\begin{equation}
\label{eq:equation7}
AEE = \frac{1}{N} \sum_{i=1}^{N} \sqrt{(v_{ix} - u_{ix})^2 + (v_{iy} - u_{iy})^2}
\end{equation}

The problem of AEE is that it is hard to get from it an intuition about the quality of velocity estimate. First of all, this error accumulates over time such that it becomes hard to compare algorithms run at different simulation frequencies. Secondly, understanding whether coding error is acceptable does not depend on the absolute value of error, but on the ratio between error and true velocity. Therefore, we used rAEE to normalize AEE by the ground truth velocity to give a relative measure of performance quality using equation \ref{eq:equation8}:

\begin{equation}
\label{eq:equation8}
rAEE = \frac{1}{N} \sum_{i=1}^{N} \frac{\sqrt{(v_{ix} - u_{ix})^2 + (v_{iy} - u_{iy})^2}}{|\mathbf{u}_i|}
\end{equation}

To compare the energy efficiency of TDE-2 and TDE-3, we also kept track of the total number of spikes emitted by the network upon presentation of the stimuli sequence.



 \section{Acknowledgments}

This work was supported by the Sony Research Award Program awarded to Prof. Guido C. H. E. de Croon.

 We want to thank Dr. Kirk Scheper for the fruitful discussions.

\section{Data Availability Statement}

The data and code that support the findings of this study are openly available in the GitHub repository at \url{https://github.com/yedutenko/TDE-3}.

---------------------------------------------------------------------------------------------------------



\end{document}